\documentclass[11pt]{article}
\usepackage[letterpaper,margin=1in]{geometry}
\usepackage{amsmath}
\usepackage{amssymb}
\usepackage{graphicx}
\usepackage[numbers]{natbib}
\usepackage[hidelinks]{hyperref}
\usepackage{url}

\usepackage{longtable}
\usepackage{booktabs}
\usepackage{siunitx}
\usepackage{longtable}

\usepackage{wrapfig}
\usepackage{xcolor}
\usepackage{placeins}
\definecolor{brown}{rgb}{0.59,0.29,0.0}

\begin{document}

\title{Correcting Performance Estimation Bias in Imbalanced Classification with Minority Subconcepts}

\author{
Taylor Maxson \and Roberto Corizzo \and Yaning Wu \and Nathalie Japkowicz \and Colin Bellinger
}

\date{}

\maketitle

\begin{abstract}
Class-level evaluation can conceal substantial performance disparities across subconcepts within the same class, causing models that perform well on average to fail on specific subpopulations. Prior work has shown that common evaluation measures for imbalanced classification are biased toward larger minority subconcepts and that utility-based reweighting using true subconcept labels can mitigate this bias; however, such labels are rarely available at test time.
We introduce a practical utility-weighted evaluation that replaces unavailable subconcept labels with predicted posterior probabilities from a multiclass subconcept model. Evaluation weights are defined as the expected utility under this posterior, yielding a soft, uncertainty-aware metric we call predicted-weighted balanced accuracy (pBA). Experiments on tabular benchmarks as well as medical-imaging and text datasets show that unweighted scores can be misleading under within-class heterogeneity, while pBA provides more stable and interpretable assessments when subconcept distributions are uneven but not pathological. Our code is available at: \url{https://anonymous.4open.science/r/correcting-bias-imbalance-9C6C/}
\end{abstract}


\section{Introduction}
\label{sec:intro}
{
The class imbalance problem has been studied for decades, yet the quality of an imbalanced classifier is still usually summarized with evaluation measures that operate at the class level \citep{Kubt1998MachineLF, Ling1998DataMF, Japkowicz2002TheCI, Prati2004ClassIV, Branco2016ASO}. This is often too coarse. In many real applications, a class is itself heterogeneous and contains multiple subconcepts with different sizes, geometries, and difficulty levels. A medical screening system, for example, may distinguish healthy from diseased patients with high average performance, while still performing poorly on rare disease subtypes. In such cases, a single class-level score may suggest reliable deployment even though some clinically meaningful subpopulations are underserved.

This paper focuses on this particular issue: \emph{performance disparity within a class}. As a running example, consider the common situation in which a practitioner reports a strong test score, say around 90\%, and reasonably concludes that any future sample should also be handled well on average. Yet this interpretation can be misleading when the target class contains multiple subconcepts. Some subpopulations may receive substantially lower performance than the  score suggests, and standard unweighted measures provide little warning that this is happening. A model can therefore look successful at the dataset level while remaining unreliable for specific groups embedded inside one class.

Prior work \citep{pmlr-v241-bellinger24a} showed that this is not merely a fairness concern but also an evaluation problem: standard measures for imbalanced classification can be biased toward the largest minority subconcepts. That study further showed that instance weighting can reduce this bias when the true subconcept labels of the test instances are known. At the same time, it revealed several lessons that motivate the present study. First, poor minority-subconcept coverage is not the whole story: sometimes the majority class is itself highly complex, so the relevant source of difficulty is distributed across the full class geometry rather than concentrated only in rare groups. Second, size alone does not determine difficulty: some small subconcepts are actually very easy, meaning that blindly distrustful treatment of all rare subconcepts is also unwarranted. Third, visualization-based analyses such as t-SNE can be useful for diagnosing when score disparity is driven by genuine overlap, isolation, or heterogeneous local structure rather than by frequency alone.

These observations suggest that an ideal performance measure should account for subconcept size without assuming that every rare subconcept is hard or that every large one is representative. In other words, the goal is not simply to replace one uniform score with another, but to construct an evaluation that reflects how performance is distributed over the latent subpopulations that make up a class. This is especially important in applications where deployment decisions are made from aggregate test scores and where underperformance on a small but important subgroup may carry disproportionate risk.

This work argues that unweighted scores can be misleading in the presence of subconcept complexity and within-class heterogeneity. To correct this, we build on previous work to develop a utility weighting strategy that only assumes that subconcept labels are available in the training data. We propose a practical utility-weighted evaluation that replaces unavailable test-time subconcept labels with predicted posterior probabilities from a
multiclass subconcept model. Evaluation weights are defined as the expected utility
\(
a_x = \sum_{s \in \mathcal{S}} q_s(x)\, u_s
\),
yielding a soft, uncertainty-aware utility weighting measure that avoids brittle hard assignments while preserving sensitivity to performance on rare but important subconcepts.

We evaluate the proposed method on real-world medical-image and multimodal hate-speech/text datasets as well as the Keel and PMLB benchmark datasets. Multiclass datasets are transformed into binary class-imbalance problems while preserving knowledge of the underlying subconcepts. Our method is compared to standard imbalanced measures and the utility weighted measure based on true subconcept labels used in the prior work. We study both whether utility weighting corrects the optimism of unweighted scores and when that correction should be expected to work. From this analysis, the following research questions are studied:


\noindent\textbf{- RQ1:} Can utility weighting based only on training-time subconcept information reduce the bias of standard unweighted measures toward larger minority subconcepts?

\noindent\textbf{- RQ2:} How does the accuracy of the subconcept predictor affect the reliability of the resulting predicted utility weights?

\noindent\textbf{- RQ3:} How do subconcept geometry, subconcept-specific difficulty, and class complexity affect when weighted and unweighted measures agree or disagree?

The remainder of the paper is organized as follows. Section \ref{sec:related} reviews related work, Section \ref{sec:setup} describes the experimental methodology, Section \ref{sec:results} presents the empirical results and their implications for evaluation under within-class heterogeneity, and Section \ref{sec:con} concludes the paper.
}

\section{Related Work}\label{sec:related}
{
This paper builds on imbalanced learning, where class skew, overlap, noise, low-density regions, and small disjuncts can all affect minority-class performance \citep{He2009LearningFI, Branco2016ASO, Lopez2013Insight, Krawczyk2016OpenChallenges, Chen2024ImbalancedSurvey}. It is most directly connected to work on within-class imbalance and small disjuncts \citep{Japkowicz2001WithinClass, Jo2004ClassIV}, and extends \citep{pmlr-v241-bellinger24a}, which showed that standard imbalanced-classification measures can be biased toward larger minority subconcepts when test-time subconcept labels are known. Related concerns appear in hidden-stratification work in medical imaging \citep{OakdenRayner2020HiddenStratification}, long-tailed learning methods that address nonuniform class frequencies \citep{Cui2019ClassBalanced, Cao2019LDAM, Ren2020BalancedMetaSoftmax, Kang2020Decoupling}, and group-robust or subpopulation-shift evaluation, where strong average performance can coexist with high error on particular groups \citep{Sagawa2019GroupDRO, Koh2021WILDS, Liu2021JustTrainTwice}. The focus here is complementary: rather than training a more robust classifier directly, training-time subconcept information is used to obtain a more informative evaluation of a fixed classifier when testing subconcept labels are unavailable.}

\section{Preliminaries}\label{sec:prelim}

We consider a binary classification problem with labels $y_i \in \{0, 1\}$, where class $1$ denotes the minority class.
Within each class, and in particular within the minority class, instances may belong to distinct \emph{subconcepts}
(e.g., semantic subgroups or latent modes) that are themselves unevenly represented.

Our goal is to evaluate classifiers in a way that: (i) treats the two classes symmetrically despite class imbalance, and
(ii) mitigates bias toward large subconcepts by increasing the influence of smaller subconcepts during evaluation. To this end, we define a utility-weighted imbalance evaluation measures. In this section, we demonstrate the utility-weighted balanced accuracy measure (UWBA).

Let $\hat{y}_i$ denote the predicted label for instance $i$. The standard balanced accuracy (BA) for binary classification is
defined as the arithmetic mean of class-wise recalls:
\begin{equation}
\label{eq:ba}
\mathrm{BA}
=
\frac{1}{2}
\left(
\mathrm{Recall}_0 + \mathrm{Recall}_1
\right), \text{where}
\end{equation}

\begin{equation}
\mathrm{Recall}_c
=
\frac{
\sum_{i=1}^n \mathbb{I}[y_i = c \wedge \hat{y}_i = c]
}{
\sum_{i=1}^n \mathbb{I}[y_i = c]
},
\quad c \in \{0,1\},
\end{equation}
and $\mathbb{I}[\cdot]$ denotes the indicator function. Balanced accuracy assigns equal importance to both classes,
regardless of class frequencies. To address imbalance among subconcepts, we associate each instance $i$ with a nonnegative utility weight $w_i$.
These weights encode the relative importance of instances during evaluation. In our setting, weights are chosen such that smaller subconcepts within a class receive higher per-instance weights, thereby counteracting dominance by larger subconcepts.

Given weights $w_i$, the weighted recall for class $c$ is defined as:
\begin{equation}
\label{eq:weighted_recall}
\mathrm{Recall}_c^{(w)}
=
\frac{
\sum_{i=1}^n w_i \, \mathbb{I}[y_i = c \wedge \hat{y}_i = c]
}{
\sum_{i=1}^n w_i \, \mathbb{I}[y_i = c]
}.
\end{equation}
\noindent
The resulting \emph{utility-weighted balanced accuracy} (UWBA) is:
\begin{equation}
\label{eq:uwba}
\mathrm{UWBA}
=
\frac{1}{2}
\left(
\mathrm{Recall}_0^{(w)} + \mathrm{Recall}_1^{(w)}
\right).
\end{equation}
\noindent
When all instance weights are identical, UWBA reduces exactly to standard balanced accuracy, ensuring backward compatibility with established practice.

\begin{figure}[t]
\centering
\includegraphics[width=0.85\linewidth]{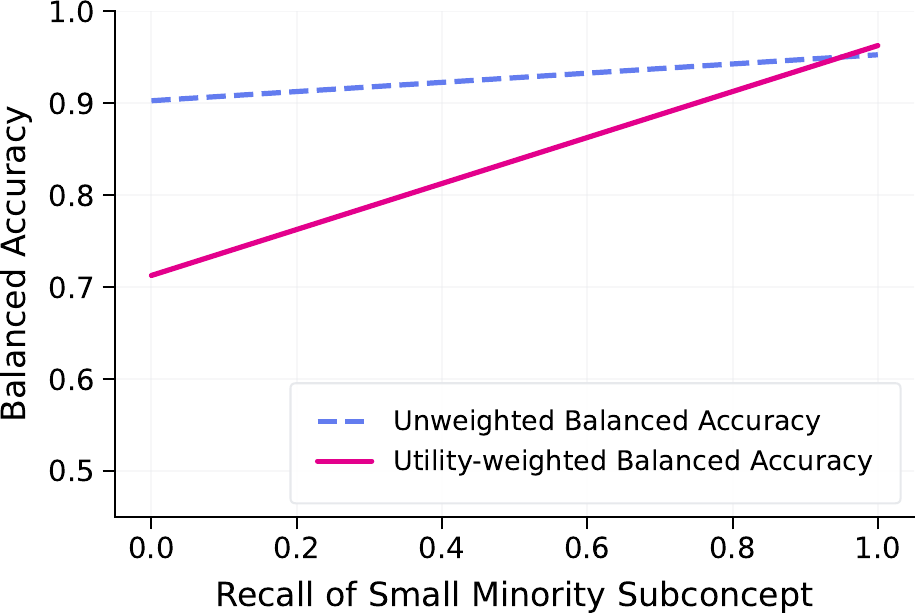}
\caption{
Balanced accuracy as a function of the recall of a small minority
subconcept. The recall of the large minority subconcept and the majority class are fixed.
Unweighted balanced accuracy (dashed) is largely insensitive to failures on the
small subconcept, while Utility-Weighted Accuracy (solid) varies
substantially, penalizing poor coverage of the small subconcept and rewarding
improvements.
}
\label{fig:uwba_sensitivity}
\end{figure}

As a simple illustration, suppose the majority recall is fixed at $0.95$ and the minority class contains two subconcepts: a large subconcept $A$ covering $90\%$ of minority instances with recall $0.95$, and a small subconcept $B$ covering $10\%$ with recall $0.10$. Standard BA uses the frequency-weighted minority recall $0.9(0.95)+0.1(0.10)=0.865$, so the severe failure on $B$ is largely hidden. If the evaluation utility gives the two minority subconcepts comparable influence, the minority recall becomes $(0.95+0.10)/2=0.525$. Figure \ref{fig:uwba_sensitivity} shows the same effect as the recall of the small subconcept varies: standard BA changes slowly because the small subconcept has little mass, whereas the utility weighted measure UWM changes more directly because the small subconcept has explicit evaluation utility. The purpose of UWM is not to assume that smaller subconcepts are always harder; it is to make the reported score sensitive to whether performance is concentrated in the largest minority subconcepts.

\section{Methodology}\label{sec:setup}
In this work, we leverage the original formulation in \citep{pmlr-v241-bellinger24a} as a starting point, and consider a more general approach to producing the utility weights, $w$, for use in the UWBA to reduce optimism in imbalance classification with heterogenous subconcepts. Below we outline our methodology.

\subsection{Datasets}
\label{subSec:Date}


The medical image experiments use four datasets. The NIH ChestX-ray dataset \citep{Wang2017ChestXray8} contains 112,120 frontal chest X-ray images from 30,805 patients with thoracic findings labels. In our experiments, we use 91,324 single-label images, covering 15 lesion types: No Finding, Infiltration, Atelectasis, Effusion, Nodule, Pneumothorax, Mass, Consolidation, Pleural\_Thickening, Cardiomegaly, Emphysema, Fibrosis, Edema, Pneumonia, and Hernia. The private Ottawa Hospital dataset contains 4,206 chest X-ray images labeled for Effusion, Pneumothorax, Subcutaneous Emphysema, and No Finding. In our experiments, we use 2,556 single-label images. The remaining two medical-image datasets are Pneumonia/COVID-19/Tuberculosis and HAM10000 \citep{Tschandl2018HAM10000}. The evaluation also includes the MMHS150K multimodal hate-speech/text dataset \citep{Gomez2020MMHS}. In addition, tabular benchmark datasets from the Keel and PMLB repositories are used \citep{Olson2017PMLB}. We select only those datasets with at least 3 subconcepts and, after downsampling, more than eight instances in the smallest subconcept. These restrictions make the endpoint comparison identifiable and avoid unstable calibration on extremely small subconcepts. The datasets are provided in Appendix Tables \ref{tab:datasets_keel8} and \ref{tab:datasets_PMLB_48}. 


Imbalanced binary classification tasks with imbalanced subconcepts are constructed from the multiclass datasets described above. For a dataset with $C$ classes, the labels are sorted from most to least frequent and at most the eight largest classes are kept, if they exist ($C' = \min(C, 8)$). These classes are then divided into a majority side and a minority side: the majority class contains the $\lfloor C'/2 \rfloor$ largest classes, while the remaining $C' - \lfloor C'/2 \rfloor$ classes form the minority class. The minority side is then down-sampled so that each successive subconcept is at least half the size of the preceding one.

\subsection{Classification, Resampling and Evaluation}
\label{subSec:classification}
The random forests (RF) classifier from the scikit-learn package is used throughout experiments. For imbalance correction, we use Random Oversampling (ROS), Random Undersampling (RUS), SMOTE, from the imblearn package, and cost-sensitive RF (weight). The results are evaluated according to the balanced accuracy (BA) measure. BA is selected for its effectiveness for performance evaluation on imbalanced data and its suitability for the utility-weighting proposed in this work. While our primary focus is on BA for its mathematics fit with the proposed utility weighting method, for completeness, we include AUC and F-measure. $5\times 2$-fold stratified cross-validation is use to achieve a robust comparison of approaches. For each tabular dataset, the procedure is repeated once for RF with each imbalance correction method used on the binary dataset. Evaluation is performed with BA and UWBA with both predicted instance weights, and the true subconcept prior weighted BA.

{The same protocol is used for the medical embeddings-plus-RF pipeline. For direct VGG16 medical classification, ten independent patient-level train/validation/calibration/test splits of 60\%/15\%/10\%/15\% are used instead, which preserves more data for representation learning and avoids patient leakage.}


\subsubsection{Subconcept Classification for Predicted Weights}

Two sources of weights are compared. The oracle, true-subconcept variant follows prior work \citep{pmlr-v241-bellinger24a}: every majority-class instance receives weight 1, while a minority instance in subconcept $s$ receives
\begin{equation}
u_{s_1}=\frac{\arg\max_{|s_0|} \in S_{0}}{|s_1|},
\end{equation}
where $|s_1|$ is the size of minority subconcept $s_1$ in the training data and $\arg\max_{|s_0|} \in S_{0}$ is the size of the largest majority subconcept. Majority instances keep weight 1, so the majority recall term remains a standard class-level recall; the correction is targeted at the minority side, where the evaluation bias of interest occurs. These true-label weighted is denoted wBA. This is the true-weight balanced-accuracy version of UWM.

The proposed practical variant replaces unavailable test-time subconcept labels with predicted membership probabilities. The multiclass subconcept classifier described above outputs $q_s(x)=\Pr(s\mid x)$ for each test instance $x$. Using the same utility value $u_s$ for each subconcept, with $u_s=1$ for majority subconcepts, the predicted evaluation weight is
\begin{equation}
a_x = \sum_{s\in\mathcal{S}} q_s(x)u_s.
\end{equation}
Thus the weight assigned to $x$ is the expected utility weight it would receive under the subconcept posterior. This soft construction is deliberate. A hard assignment would force each test instance into one predicted subconcept and would make the evaluation sensitive to arbitrary classifier mistakes near subconcept boundaries. The posterior-weighted version instead preserves uncertainty: if an instance is plausibly split across a large and a small subconcept, its evaluation weight is intermediate. The predicted-weighted measure is denoted pBA; pBA is the predicted-weight balanced-accuracy version of UWM.

\subsection{Assessment of Evaluation Bias}
{
The binary models are trained on the full training set defined by the relevant split; for evaluation, however, performance statistics are obtained for multiple subsets of the testing data. {This permits evaluation of a single deployed classifier at both aggregate and endpoint-subconcept levels, rather than training separate models for each subgroup.}

The full testing data is denoted $X_{\text{tst}}^{\text{full}}$, and the full majority testing class is denoted $X_{\text{tst}}^{0}$. The $i$-th minority class from the testing data is denoted $X_{\text{tst},i}^{1}$, where $|S|$ is the number of minority subconcepts and $i\in[|S|]$; the minority subconcepts are assumed to be ordered from largest to smallest. Then define $X_{\text{tst}}^{\text{sub}_i} = X^{0}_{\text{tst}} \cup X^{1}_{\text{tst},i}$ and, for notation, $X^{\text{largest}}_{\text{tst}} = X^{\text{sub}_1}_{\text{tst}}$ and $X^{\text{smallest}}_{\text{tst}} = X^{\text{sub}_{|S|}}_{\text{tst}}$. Performance statistics are extracted from $X^{\text{full}}_{\text{tst}}$, $X^{\text{largest}}_{\text{tst}}$, and $X^{\text{smallest}}_{\text{tst}}$ for every dataset. 

The central assessment asks whether an evaluation measure on $X^{\text{full}}_{\text{tst}}$ behaves more like the same measure on $X^{\text{largest}}_{\text{tst}}$ than on $X^{\text{smallest}}_{\text{tst}}$. For each measure $M$, the Pearson correlations $M(X^{\text{full}}_{\text{tst}})$--$M(X^{\text{largest}}_{\text{tst}})$ and $M(X^{\text{full}}_{\text{tst}})$--$M(X^{\text{smallest}}_{\text{tst}})$ are compared after averaging fold-level scores into one dataset-level estimate. A useful weighting scheme should reduce the magnitude of this correlation difference; the sign only identifies which endpoint is more aligned with the full-test score. Positive bars in the correlation-difference figures indicate stronger alignment with the largest minority subconcept, bars near zero indicate a more balanced relationship, and negative bars indicate possible overcorrection toward the smallest subconcept.}

\subsection{Subconcept Classifier Evaluation}\label{subSec:subconcept_classifier_eval}

{To assess whether the subconcept classifier is good enough to support the predicted weighting scheme, subconcept-classifier performance is compared with the gap between the true and predicted weights. For a subconcept $C$, let $\mathrm{true}_j(i)$ denote the true weight assigned to instance $i$ in the $j$th cross-validation iteration, let $\mathrm{predicted}_j(i)$ denote the corresponding predicted weight, and let $C_j$ denote the testing instances in subconcept $C$ during that iteration. The true and predicted weights for $C$ are summarized across the ten test folds as
\begin{align*}
    \bar{t}_C &=
    \frac{\sum_{j\in [10]} \sum_{i\in C_j} \mathrm{true}_j (i)}
    {\sum_{j\in[10]}|C_j|},\\
    \bar{p}_C &=
    \frac{\sum_{j\in [10]} \sum_{i\in C_j} \mathrm{predicted}_j (i)}
    {\sum_{j\in[10]}|C_j|}.
\end{align*}
Letting $S$ denote the set of subconcepts, the average absolute true--predicted weight gap is computed as
\begin{equation}
    \mu = \frac{\sum_{C\in S} \left|\bar{t}_C - \bar{p}_C\right|}{|S|}.
\end{equation}
Finally, the Pearson correlation between $\mu$ and the average subconcept-classifier performance across the ten iterations is calculated, removing outlier datasets whose $\mu$ value or average classifier performance has a z-score greater than 4. This analysis is separate from the largest-versus-smallest correlation-gap analysis used for RQ1. Here, $\mu$ measures only how far the predicted weights are from the true weights; it is not the correlation gap between $X^{\text{largest}}_{\text{tst}}$ and $X^{\text{smallest}}_{\text{tst}}$. The practical question is therefore simple: when the subconcept classifier is better, do the predicted weights become closer to the true weights? A negative correlation answers yes, because higher subconcept-classifier performance is associated with a smaller true--predicted weight error. A positive correlation would answer no, because it would mean that better subconcept classification is associated with larger weight error.}
 

\section{Experimental Results}\label{sec:results}

This section answers the three research questions.

\subsection{RQ1: Can instance weighting the evaluation measure using only the subconcept labels of the testing instances correct the bias towards the larger minority subconcept(s)?}
\subsubsection{Keel Datasets}
 The results of this analysis for the instance weighted measures are shown in the last six columns of Table \ref{tab:corr_t_w_p} (pAUC, pBA, pF1, wAUC, wBA, wF1). The results using the standard measures are shown in the ``Unweighted'' columns. The table shows that, without using any imbalance correction methods, both the predicted weights and the weights based on the true subconcepts narrow the gap between the performance estimate correlations, though they do not close them.


{The main pattern in Table \ref{tab:corr_t_w_p} is that predicted weighting reduces the largest-versus-smallest correlation gap for balanced accuracy and F1, while true weighting gives the strongest correction. For AUC, predicted weighting also reduces the Keel gap, but true weighting can slightly overshoot; AUC is therefore treated separately from BA and F1.}

\begin{table*}[!t]

\centering
\scriptsize
\setlength{\tabcolsep}{4pt}
\caption{Pearson correlations between full-test performance and largest/smallest-subconcept performance on the Keel datasets, with no imbalance correction.}
\label{tab:corr_t_w_p}
\resizebox{\textwidth}{!}{
\begin{tabular}{llccccccccc}
     \toprule
     {} & {} & \multicolumn{3}{c}{Unweighted} & \multicolumn{3}{c}{Predicted Weights} & \multicolumn{3}{c}{True Weights}\\
     \cmidrule{3-5}\cmidrule{6-8}\cmidrule{9-11}
     Subconcept & & {\bf AUC}    & {\bf BA}  & {\bf F1}   & {\bf pAUC}    & {\bf pBA}  & {\bf pF1} &{\bf wAUC}&{\bf wBA}&{\bf wF1}\\
     \midrule
     {\bf Largest} & Correlation
& 0.885  &   0.946 & 0.949 
& 0.824 &   0.789 & 0.814 
&0.853  &0.708  &0.732 \\
& p-value &
3.49e-3 & 3.78e-4 & 3.19e-4 & 
1.18e-2 & 0.020 & 0.014
& 3.96e-2 & 0.049 & 0.039\\
{\bf Smallest} & Correlation
& 0.670 & 0.194 & 0.205 
& 0.737  & 0.442  & 0.435 
&0.853 & 0.590 & 0.552\\
&p-value &
6.94e-2 & 0.645 & 0.626 &
3.71e-2 & 0.273 & 0.281
&7.07e-3 & 0.124 & 0.156\\
\bottomrule
\end{tabular}}
\end{table*}

\begin{figure*}[!t]
    \centering
    \includegraphics[width=0.88\linewidth]{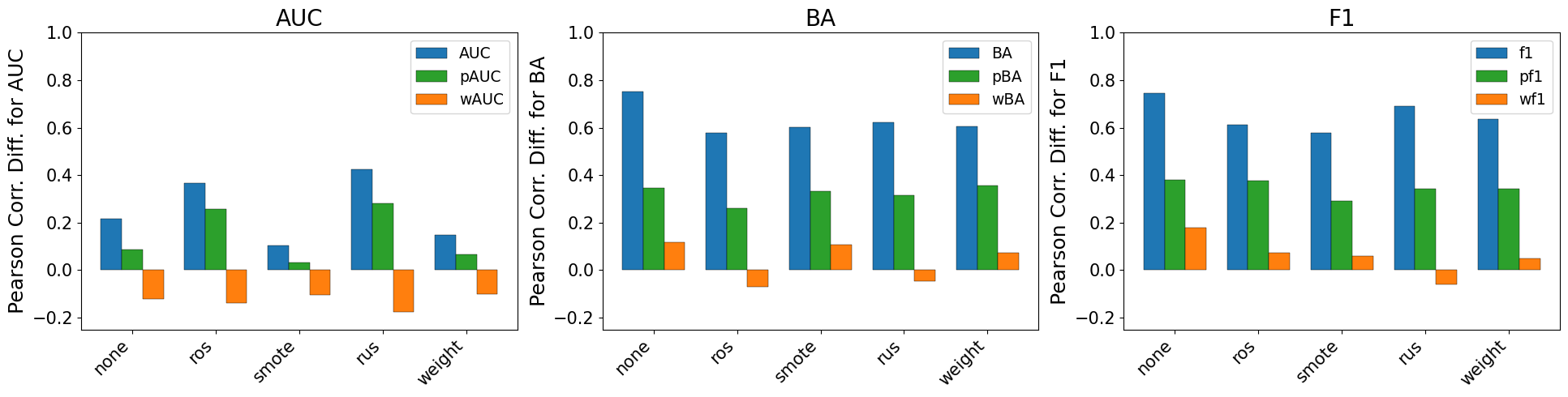}
    \caption{Keel correlation-gap results: difference between the full--largest and full--smallest Pearson correlations for AUC, pAUC, and wAUC (left); BA, pBA, and wBA (centre); and F1, pF1, and wF1 (right), after imbalance correction.}
    \label{fig:imb_cor_diff_8}
\end{figure*}


\subsubsection{PMLB Datasets}
{The PMLB results, summarized in Table \ref{tab:corr_t_w_p_PMLB} and Appendix Figure \ref{fig:imb_cor_diff_PMLB}, provide the strongest evidence for the proposed method. For balanced accuracy and F-measure, predicted weighting reduces the correlation gap relative to the unweighted scores, and true weighting reduces it further. The correction is not complete, but it moves the full-test estimate away from being dominated by the largest minority subconcept while requiring only training-time subconcept labels.}

{AUC is less conclusive. On PMLB, the unweighted AUC gap is already much smaller than the BA and F1 gaps, and predicted weighting does not consistently improve it. Thus, the answer to RQ1 is strongest for BA and F1: predicted weighting is a practical compromise that approximates the unavailable true-weight correction and reduces the tendency of the full-test score to track the largest minority subconcept.}

\begin{table*}[!t]
\centering
\setlength{\tabcolsep}{4pt}
\caption{Pearson correlations between full-test performance and largest/smallest-subconcept performance on the PMLB datasets, with no imbalance correction.}
\label{tab:corr_t_w_p_PMLB}
\resizebox{\textwidth}{!}{
\begin{tabular}{llccccccccc}
     \toprule
     {} & {} & \multicolumn{3}{c}{Unweighted} & \multicolumn{3}{c}{Predicted Weights} & \multicolumn{3}{c}{True Weights}\\
     \cmidrule{3-5}\cmidrule{6-8}\cmidrule{9-11}
     Subconcept & & {\bf AUC} & {\bf BA} & {\bf F1} & {\bf pAUC} & {\bf pBA} & {\bf pF1} & {\bf wAUC} & {\bf wBA} & {\bf wF1}\\
     \midrule
{\bf Largest} & Correlation
& 0.995 & 0.985 & 0.988
& 0.983 & 0.955 & 0.956
& 0.981 & 0.936 & 0.947\\
& p-value
& 3.75e-47 & 1.05e-36 & 1.14e-38
& 1.62e-35 & 6.94e-26 & 3.60e-26
& 2.99e-34 & 1.93e-22 & 2.29e-24\\
{\bf Smallest} & Correlation
& 0.913 & 0.789 & 0.791
& 0.889 & 0.826 & 0.809
& 0.946 & 0.889 & 0.875\\
& p-value
& 1.56e-19 & 2.80e-11 & 2.28e-11
& 3.52e-17 & 4.89e-13 & 3.41e-12
& 3.22e-24 & 3.37e-17 & 4.34e-16\\
\bottomrule
\end{tabular}}
\end{table*}

{Table \ref{tab:corr_t_w_p_PMLB} makes the scale of the effect clearer. For BA, the full-test score has a correlation of $0.985$ with the largest-subconcept score but only $0.789$ with the smallest-subconcept score. Predicted weighting narrows this gap from $0.196$ to $0.129$, and true weighting narrows it further to $0.047$. F1 shows the same pattern. AUC starts with a smaller gap, so there is less room for improvement and the true-weighted version shifts the score closer to the smallest-subconcept endpoint. This supports the interpretation that predicted weights recover part, but not all, of the correction that would be obtained if true test-time subconcept labels were available.}

\subsection{RQ2: How does the performance of the subconcept classifier (used to correct the bias) affect the predicted weights?}\label{subSec:RQTwo}
This section reports whether better subconcept classification is associated with predicted weights that are closer to the true weights, using the methodology in Section \ref{subSec:subconcept_classifier_eval}.



{The predicted weights behave as desired: when the subconcept classifier is better, the predicted weights tend to be closer to the true weights. The correlations are negative for all three measures on both benchmark collections: $-0.868$, $-0.891$, and $-0.861$ on Keel, and $-0.300$, $-0.393$, and $-0.241$ on PMLB, for ROC AUC, balanced accuracy, and F-measure, respectively. Thus, the proposed method is most reliable when the subconcept classifier is reliable, and improving subconcept classification is a direct way to improve the predicted weights.}

\subsection{RQ3: How does individual subconcept performance affect the weighted measures?}
{Here, model accuracy refers to the final binary classifier's accuracy on each subconcept, not to the subconcept classifier used to predict weights. RQ2 asks whether the subconcept classifier produces accurate weights; RQ3 asks how the final classifier's uneven performance across subconcepts changes the weighted score. In this case, per-subconcept accuracies are used only after evaluation to explain why weighted and unweighted scores differ.}

\begin{figure}[!t]
    \centering
    \includegraphics[width=\linewidth]{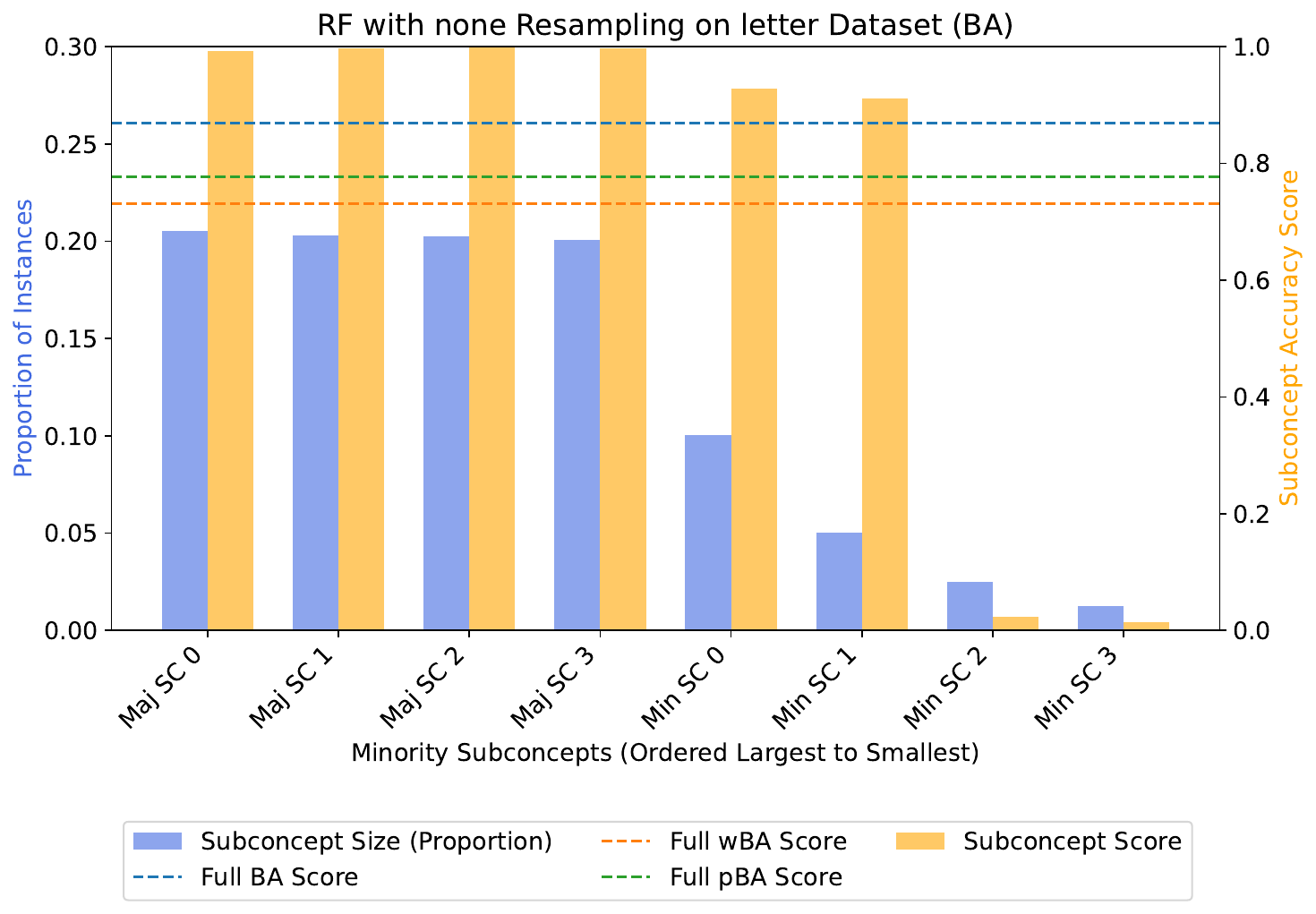}
    \vspace{0.25em}
    \includegraphics[width=\linewidth]{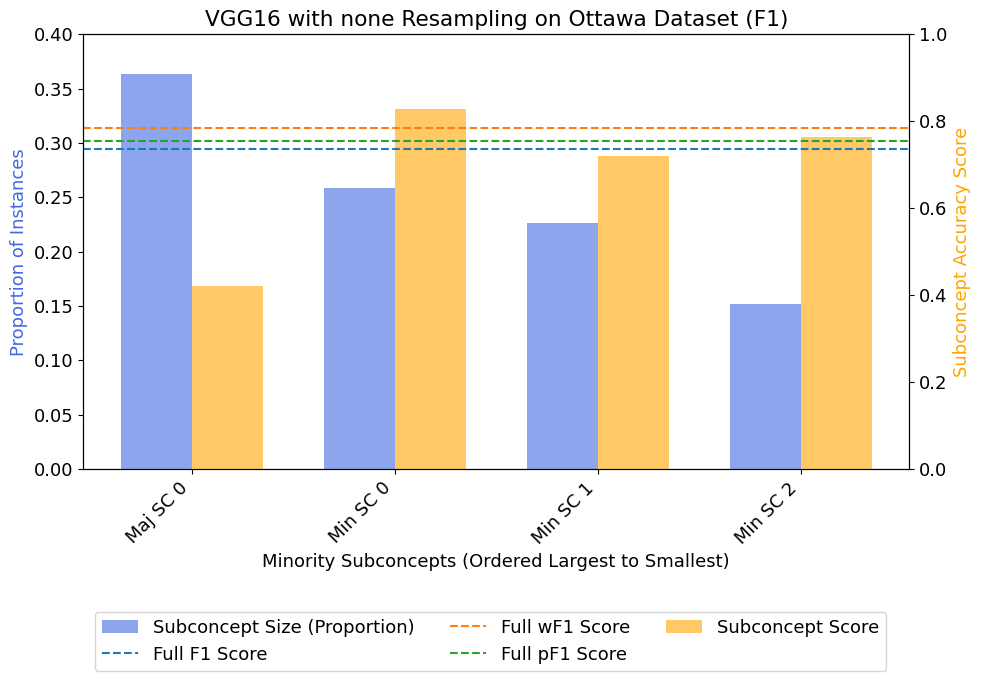}
    \caption{Representative tabular and medical-image cases. Top: PMLB balanced accuracy on \texttt{letter}, where predicted weighting moves the full-test score toward the true weighted score. Bottom: Ottawa direct VGG16 F1 counterexample, where smaller subconcepts are easier and reweighting can increase the reported score.}
    \label{fig:representative_cases}
\end{figure}

\begin{figure}[!t]
    \centering
    \includegraphics[width=\linewidth]{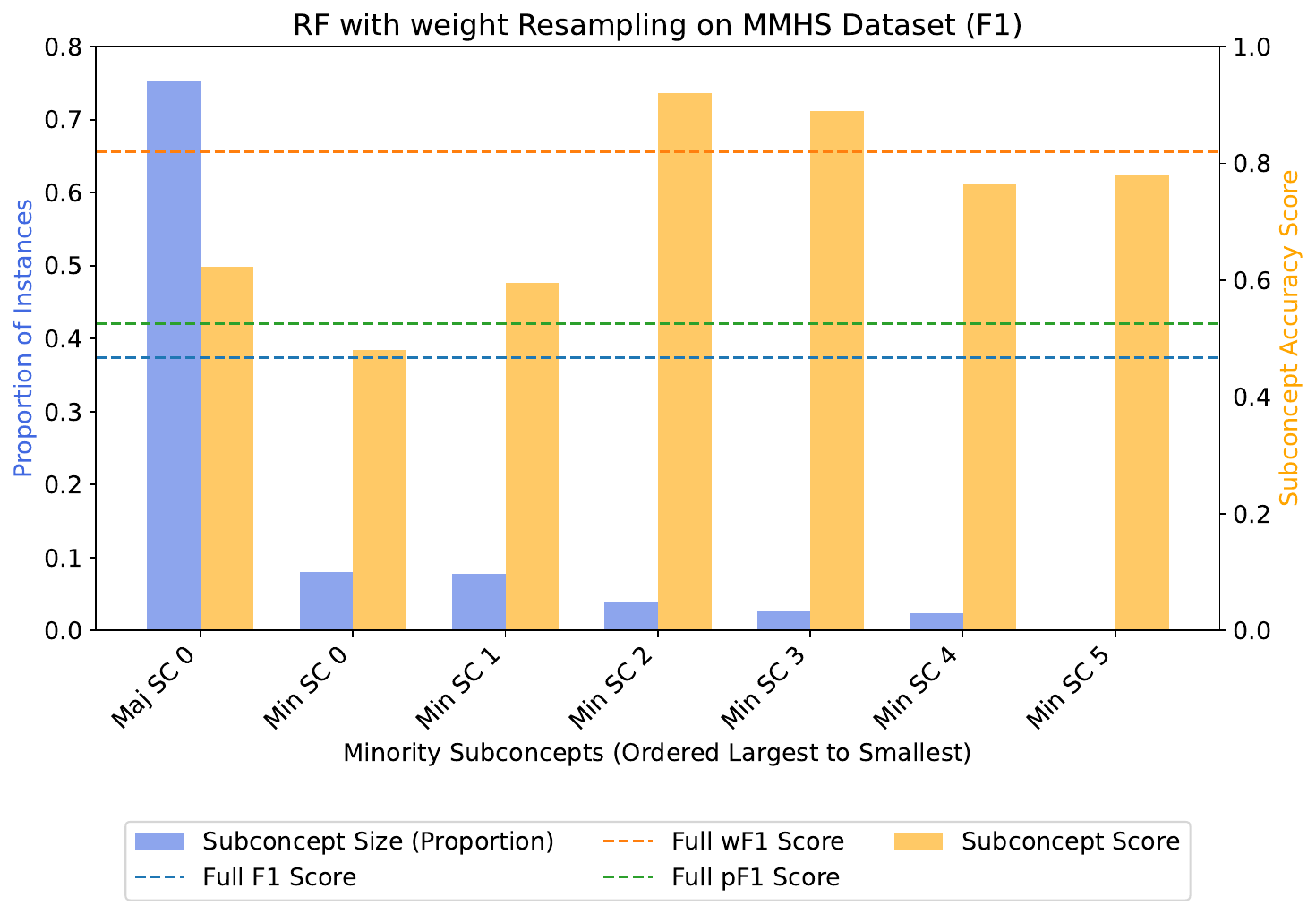}
    \caption{Representative MMHS150K text-domain counterexample with cost-sensitive RF, showing that reweighting can increase the reported score when smaller subconcepts are easier.}
    \label{fig:representative_cases_mmhs}
\end{figure}

{Figure \ref{fig:keel_ba_subconcept_plots} and the additional PMLB, text-domain, and medical-domain plots in Appendix Figures \ref{fig:imb_cor_diff_PMLB}--\ref{fig:medical_vgg_f1} show that the gap between weighted and unweighted evaluation is driven by the interaction between subconcept size and subconcept difficulty. Figures \ref{fig:representative_cases} and \ref{fig:representative_cases_mmhs} give representative cases: reweighting can lower an optimistic score when smaller subconcepts are harder, but it can also raise the score when those subconcepts are easier. This is the intended behavior: the method gives underrepresented subconcepts more influence rather than assuming that all rare subconcepts are difficult.}

\begin{figure*}[!t]
    \centering
    \includegraphics[width=\textwidth]{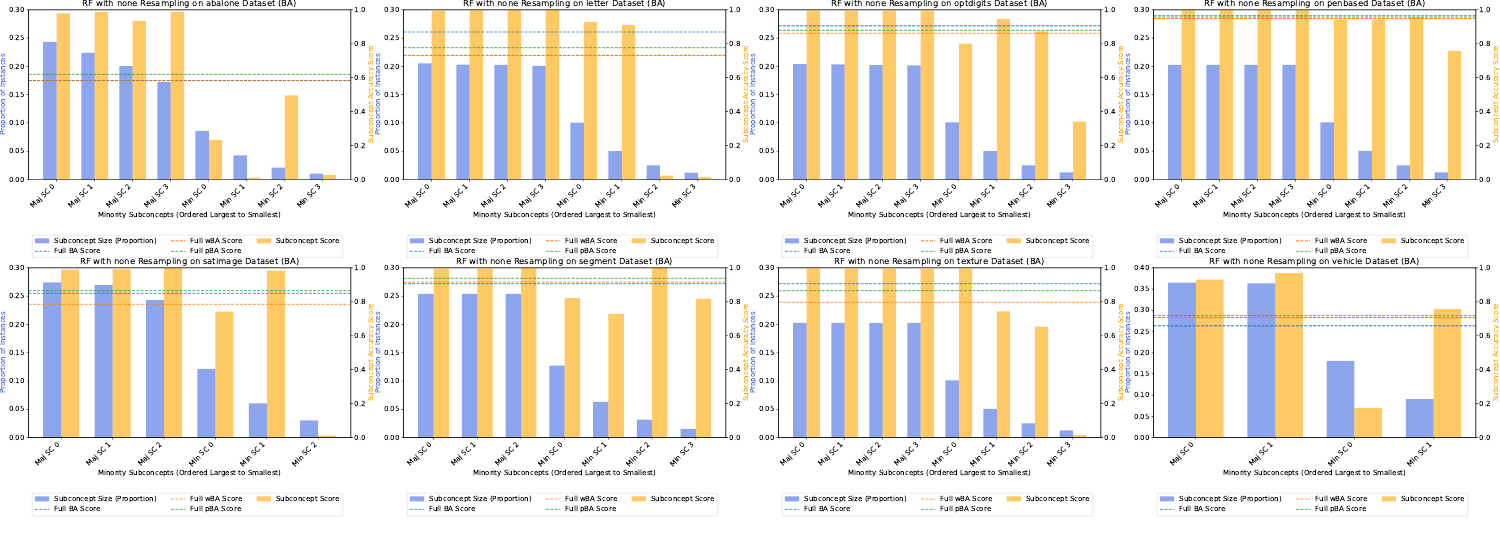}
    \caption{Balanced-accuracy comparison plots on the Keel tabular datasets. The solid blue bars show subconcept size, the yellow bars show subconcept performance, the dashed blue line shows the unweighted score, the green line shows the predicted-weight score, and the orange line shows the true-weight score.}
    \label{fig:keel_ba_subconcept_plots}
\end{figure*}

{These cases clarify how the weighted scores should be interpreted. The proposed weights do not assume that rare subconcepts are harder, and they are not intended to make every reported score smaller. Rather, they change the estimand from an instance-average score, which is dominated by large subconcepts, toward a score in which minority subconcepts have more comparable influence. When smaller subconcepts are harder, the predicted-weighted score can expose optimism in the unweighted estimate; when smaller subconcepts are easier, it can move upward. This is not a failure of the weighting scheme, but evidence that weighted evaluation and imbalance-corrected training solve different problems: rebalancing can help the classifier learn, whereas predicted subconcept weighting makes the reported performance less dependent on the accidental composition of the test set.}

{The appendix plots support the same interpretation across the cases omitted from the main text. The representative figures are therefore not intended as isolated examples, but as compact illustrations of the two behaviors that appear repeatedly in the full results: predicted weighting moves the reported score toward the true subconcept-weighted score when the test distribution would otherwise overemphasize large, easy subconcepts, and it does not force a pessimistic correction when smaller subconcepts are comparatively easy. This is the practical reason for reporting both the correlation-gap analysis and the subconcept-level plots. The former gives an aggregate test of whether full-test performance tracks the largest subconcept too strongly, while the latter explains why the weighted score moves up or down on a particular dataset.}

The medical-image and text-domain experiments are important for this interpretation because they move the analysis beyond tabular benchmark artifacts. In these domains, subconcepts correspond to meaningful sources of heterogeneity, such as lesion categories, disease labels, or attack types, and the operational concern is not only whether the average classifier score is high, but whether that score is representative of the subpopulations on which the system will be used. The same qualitative behavior across tabular, medical, and text datasets suggests that the issue is not tied to a particular feature representation or classifier family. Instead, it is a general evaluation problem: whenever one class is internally heterogeneous, the class-level score can be disproportionately determined by the largest or easiest subconcept.

The results also indicate when the proposed evaluation should be interpreted cautiously. Predicted weighting is useful only to the extent that the training data contain informative subconcept structure and the subconcept classifier can recover that structure on held-out examples. When subconcept prediction is weak, the predicted weights can only approximate the desired true-subconcept correction, and the unweighted, predicted-weighted, and true-weighted scores should be read together rather than as interchangeable summaries. This is why the paper reports both aggregate correlations and dataset-level plots: the correlations show the systematic bias pattern, while the plots expose the local conditions under which the correction is reliable, incomplete, or unnecessary.

Taken together, the results suggest a practical reporting strategy for imbalanced problems with known or discoverable subconcepts. Standard BA, F1, and AUC remain useful summaries, but they should be accompanied by a subconcept-sensitive estimate when the minority class is internally heterogeneous. The weighted score is most informative when it disagrees with the unweighted score: a lower predicted-weighted score warns that the aggregate result is optimistic for smaller or harder subconcepts, while a higher predicted-weighted score indicates that rare subconcepts are not necessarily the source of the performance problem. The comparison is therefore diagnostic rather than purely corrective; it tells the reader which part of the test distribution is driving the headline score.

The strongest evidence for this interpretation comes from balanced accuracy and F-measure. For these measures, the benchmark results show the expected ordering: unweighted scores are most aligned with the largest minority subconcepts, true subconcept weights reduce this dependence most strongly, and predicted weights usually fall between the two. The AUC results are more cautious because AUC is ranking-based rather than an accuracy-style measure; predicted weighting can still expose sensitivity to subconcept composition, but the clearest evidence is for BA and F1.

The representative plots and Pearson correlation tables are therefore complementary. The tables summarize whether full-test performance systematically tracks the largest or smallest subconcept, while the plots explain individual cases by showing each subconcept's size and difficulty. Across tabular, medical-image, and text-domain datasets, the same lesson appears: the method is most appropriate as a reporting layer for applications where class labels are too coarse to describe the populations of interest. It asks whether the headline score is robust to within-class heterogeneity, not whether every small subconcept has been fixed by the classifier.

\section{Conclusion and Future Work}\label{sec:con}
This work introduces a predicted-subconcept weighting technique for reducing bias in standard imbalanced-classification measures when minority classes contain heterogeneous subconcepts. The method requires subconcept labels only in the training data, making it usable when test-time subconcept membership is unavailable. Across tabular, medical-image, and text-domain experiments, unweighted scores can hide which subconcepts drive performance, while predicted weighting gives a more explicit estimate under within-class heterogeneity.

Predicted weighting is an evaluation procedure, not a substitute for training-time imbalance correction or for improving the classifier itself. Its purpose is to make reported performance less sensitive to the accidental mixture of subconcepts in the test set, which matters in applications such as medical imaging and harmful-content detection where high average scores can coexist with subgroup failures.

The most useful role of the proposed score is therefore comparative. When the predicted-weighted and unweighted scores are similar, the standard aggregate score is less likely to be dominated by a particular subconcept. When they differ, the disagreement identifies a case where deployment decisions should not rely on the unweighted score alone. This makes the method a lightweight diagnostic for deciding when more detailed subgroup analysis is needed.

This framing is especially important because the method does not require new assumptions about which subconcepts should be difficult. Instead, it uses the observed training subconcept structure to make the test estimate less dependent on the realized test mixture. The resulting score is not meant to replace domain-specific auditing, but to flag when the usual class-level summary is too coarse to support a deployment decision.

Future work should improve the subconcept-prediction step, test additional base classifiers, examine whether similar evaluation bias arises for large majority subconcepts, and replace the endpoint largest-versus-smallest analysis with a distributional view over all subconcepts.

\clearpage

\bibliographystyle{siamplain}

\appendix
\clearpage

\section{Additional results}

This appendix provides the supporting tables and plots for the experimental results summarized in the main paper. The main paper keeps only the results needed to establish the central argument: unweighted class-level scores can be dominated by large or easy minority subconcepts, and predicted subconcept weighting can move the reported score toward the score that would be obtained if true test-time subconcept labels were available. The appendix gives the fuller empirical record behind that claim.

The appendix is organized in the same order as the experimental questions in Section \ref{sec:results}. Tables \ref{tab:datasets_keel8} and \ref{tab:datasets_PMLB_48} first document the tabular benchmark construction. The original Keel and PMLB problems are multiclass datasets; for the experiments, groups of original labels are merged to form a majority class and a minority class, while the original labels inside the minority class define the subconcepts. These tables are included so that the engineered binary tasks are explicit and reproducible, and so that the reader can see the level of class imbalance present before considering within-class subconcept imbalance.

\begin{figure*}[p]
    \centering
    \fcolorbox{red}{white}{\includegraphics[width=0.98\linewidth]{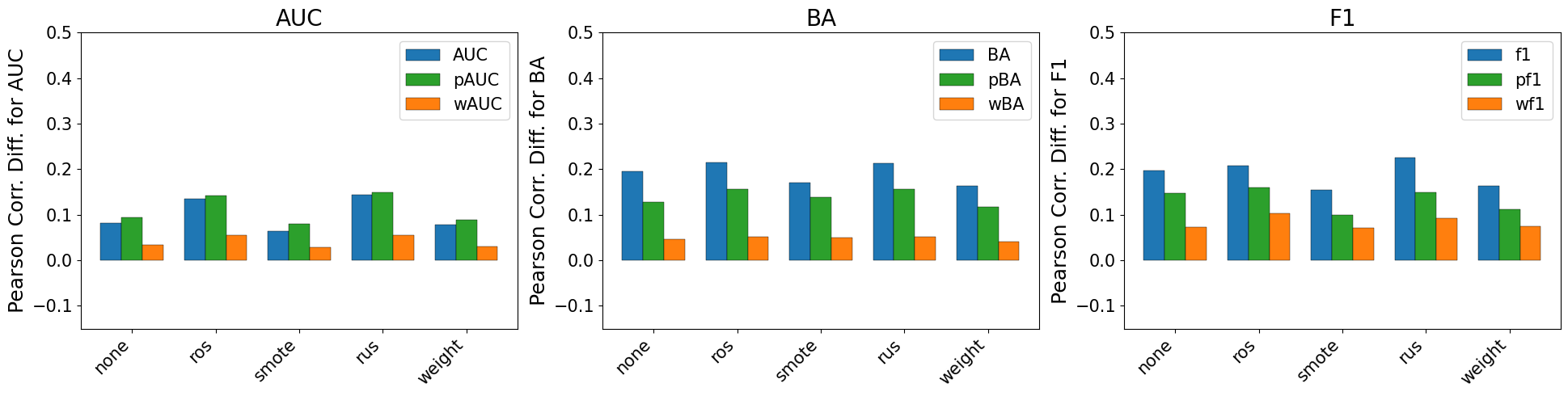}}
    \caption{The difference between the Pearson correlations on $X^{\text{full}}_{\text{tst}}$ and $X^{\text{largest}}_{\text{tst}}$ and on $X^{\text{full}}_{\text{tst}}$ and $X^{\text{smallest}}_{\text{tst}}$ for AUC, pAUC, and wAUC (left); BA, pBA, and wBA (centre); and F1, pF1, and wF1 (right) after the application of the imbalance correction methods on the PMLB datasets in Table \ref{tab:datasets_PMLB_48}.}
    \label{fig:imb_cor_diff_PMLB}
\end{figure*}

Figure \ref{fig:imb_cor_diff_PMLB} extends the RQ1 correlation-gap analysis to the full PMLB collection. The quantity plotted is the difference between how strongly full-test performance correlates with the largest minority subconcept and how strongly it correlates with the smallest minority subconcept. Large positive differences indicate that the full-test score is tracking the largest subconcept more closely than the smallest one. The practical reading is therefore based on magnitude and direction: predicted weighting is useful when it reduces this gap relative to the unweighted measure, and true weighting provides the reference behavior that would be possible if test-time subconcept labels were known.

The remaining tabular figures report the dataset-level behavior behind the aggregate correlations. For each dataset, the blue bars show subconcept prevalence, the yellow bars show the classifier's performance on each subconcept, and the horizontal lines show the unweighted, predicted-weighted, and true-weighted full-test scores. These plots should be read as diagnostic explanations rather than as independent hypothesis tests. They show whether a weighted score moves downward because small subconcepts are harder, moves upward because small subconcepts are easier, or remains close to the unweighted score because subconcept difficulty is relatively uniform.

The MMHS150K and medical-image figures serve the same purpose in less controlled domains. They show that the qualitative issue is not limited to tabular benchmark construction: when classes contain meaningful internal groups, the reported average can depend strongly on which groups are common in the test set. The appendix includes all metric variants and imbalance-correction settings so that the representative examples in the main paper can be checked against the broader result set. Across these figures, the intended comparison is not whether predicted weighting always lowers performance, but whether it makes the reported performance less dependent on the accidental mixture of subconcept sizes and difficulties.

{
\begin{table*}[p]

\begin{center}
\setlength{\tabcolsep}{4pt}
\begin{tabular}{||l|l|l|l|l|l||}
\hline
{\bf Data Set}        & {\bf Maj. }  & {\bf Min. } & {\bf Maj.Class} & {\bf Min.Class} & {\bf I.R.} \\ \hline\hline
Abalone  & 2,377 & 454 & 8, 9, 10, 11         &  6, 7, 12, 13 & 5.24\\\hline
Letter& 3,217 & 745 & D, P, T, U& A, M, X, Y & 4.32\\\hline
Optdigits & 2,277&529& 1, 3, 4, 7& 2, 5, 6, 9& 4.30\\\hline
Penbased & 4,574 & 1,069 & 0, 1, 2, 4 & 5, 6, 7, 8 & 4.28\\ \hline
Satimage & 4,399 & 1,187 & 1, 3, 7 & 2, 4, 5 & 3.71\\ \hline
Segment & 990 & 308 & 5, 6, 7 & 1, 2, 3, 4 & 3.21\\ \hline
Texture & 2,000 & 468 & 6, 7, 8, 9 & 2, 3, 4, 10 & 4.27\\ \hline
Vehicle&435&162&bus, saab&opel, van&2.69\\\hline
\hline
\end{tabular}
\end{center}
\caption{This table indicates the multiclass datasets selected from the Keel repository for the benchmark experiments and shows the way in which they were transformed into binary classification problems.}\label{tab:datasets_keel8}
\end{table*}
}

{
\begin{table*}[p]

\centering
\scriptsize
\setlength{\tabcolsep}{4pt}
\begin{tabular}{llll}
\toprule
\multicolumn{4}{c}{PMLB datasets used in the benchmark experiments}\\
\midrule
allbp & allhypo & allrep & analcatdata authorship \\
analcatdata dmft & analcatdata germangss & ann thyroid & balance scale \\
calendarDOW & car evaluation & cars & connect 4 \\
dna & ecoli & fars & hayes roth \\
iris & kddcup & krkopt & led24 \\
led7 & letter & mfeat factors & mfeat fourier \\
mfeat karhunen & mfeat morphological & mfeat pixel & mfeat zernike \\
mnist & new thyroid & nursery & optdigits \\
page blocks & pendigits & penguins & poker \\
satimage & schizo & segmentation & sleep \\
splice & tae & texture & vehicle \\
waveform 21 & waveform 40 & wine quality red & wine recognition \\
\bottomrule
\end{tabular}
\caption{PMLB datasets selected for the benchmark experiments. Each dataset was transformed into an engineered binary problem using the construction described in Section \ref{subSec:Date}.}\label{tab:datasets_PMLB_48}
\end{table*}
}


\begin{figure*}[p]
    \centering
    \includegraphics[page=1,width=0.98\textwidth]{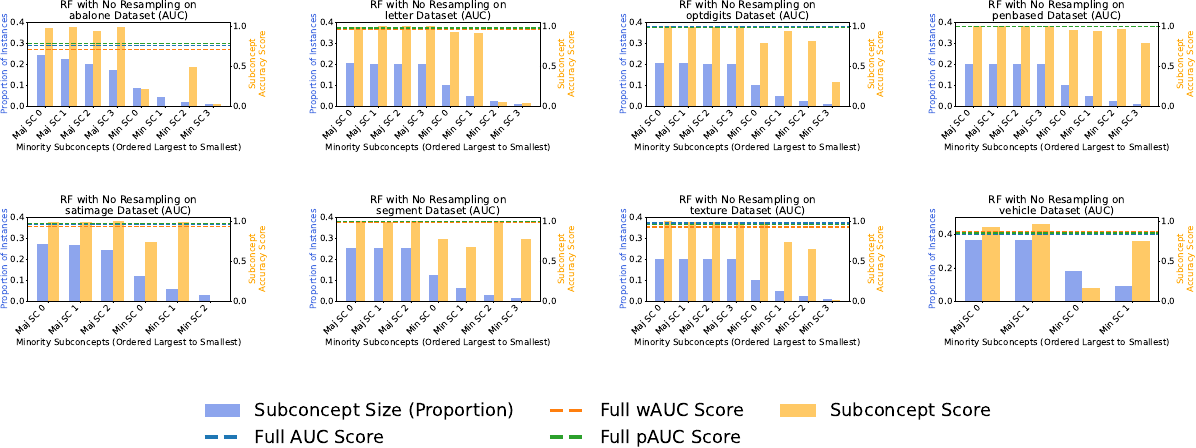}
    \caption{ROC-AUC comparison plots on the Keel tabular datasets. The interpretation of bars and lines is the same as in Figure \ref{fig:keel_ba_subconcept_plots}.}
    \label{fig:keel_auc_subconcept_plots}
\end{figure*}


\begin{figure*}[p]
    \centering
    \includegraphics[page=1,width=0.98\textwidth]{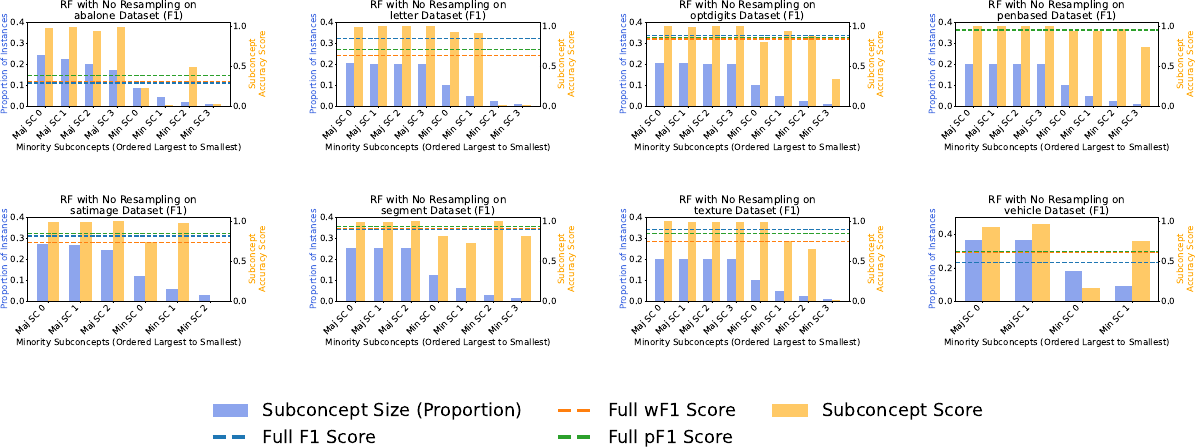}
    \caption{F1 comparison plots on the Keel tabular datasets. The interpretation of bars and lines is the same as in Figure \ref{fig:keel_ba_subconcept_plots}.}
    \label{fig:keel_f1_subconcept_plots}
\end{figure*}


\begin{figure*}[p]
    \centering
    \includegraphics[page=1,width=0.89\textwidth]{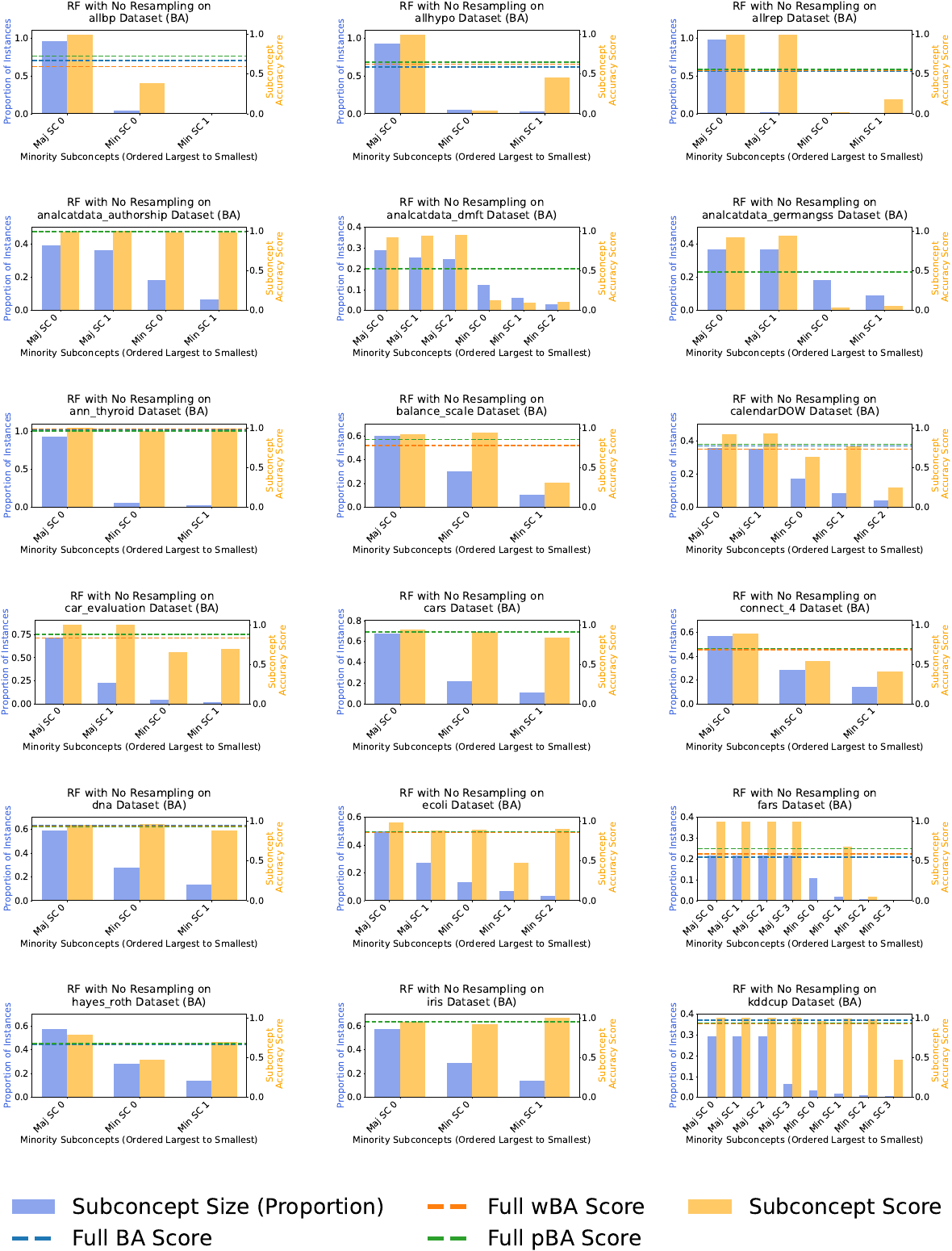}
    \caption{Balanced-accuracy comparison plots on the 48 PMLB datasets (part 1 of 3). The solid blue bars show subconcept size, the yellow bars show subconcept performance, the dashed blue line shows the unweighted score, the green line shows the predicted-weight score, and the orange line shows the true-weight score. Datasets are ordered alphabetically across the contact sheets.}
    \label{fig:pmlb_ba_subconcept_plots}
\end{figure*}

\begin{figure*}[p]
    \centering
    \includegraphics[page=2,width=0.89\textwidth]{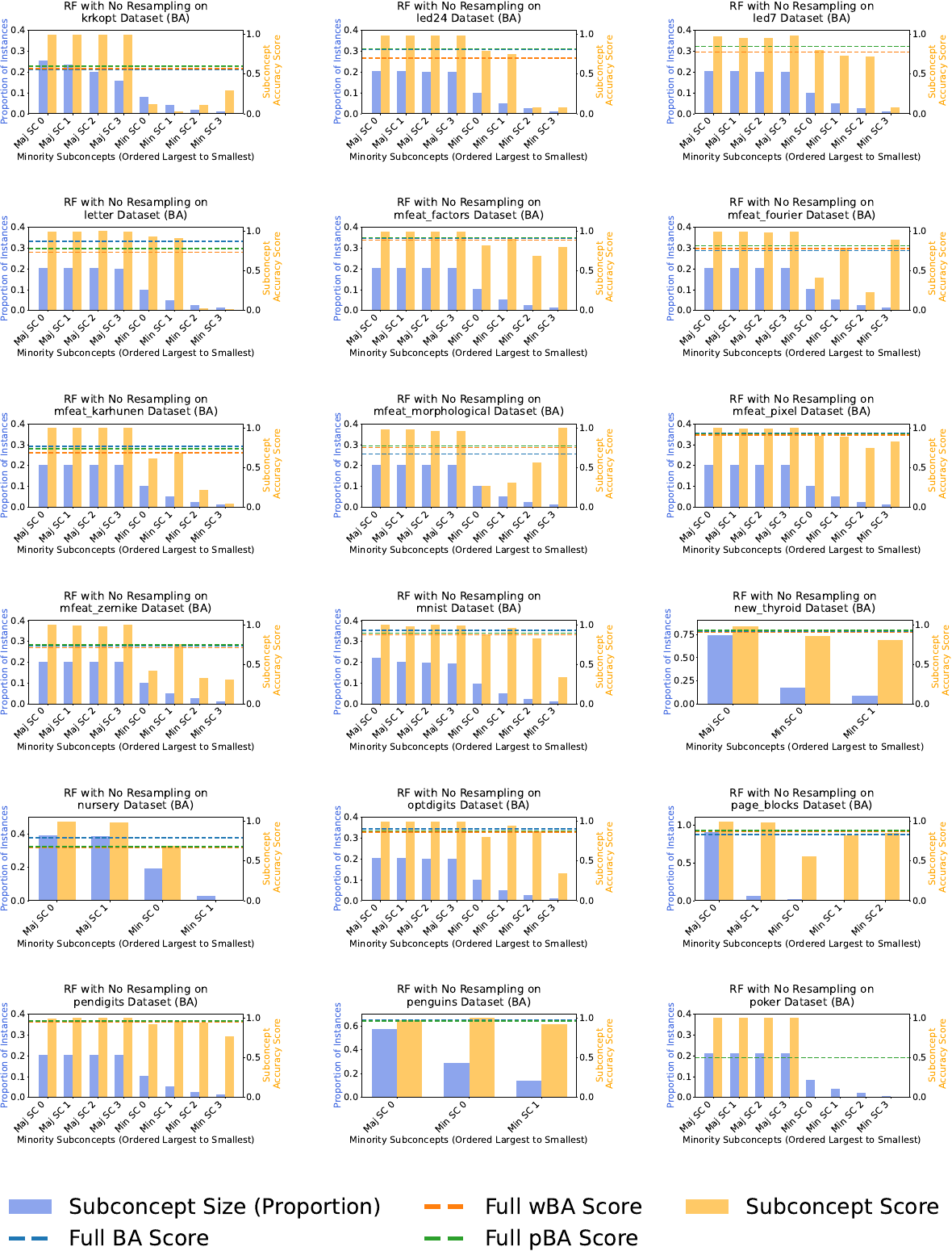}
    \caption{BA comparison plots on the 48 PMLB datasets (part 2 of 3). The interpretation of bars and lines is the same as in Figure \ref{fig:pmlb_ba_subconcept_plots}.}
    \label{fig:pmlb_ba_subconcept_plots_2}
\end{figure*}

\begin{figure*}[p]
    \centering
    \includegraphics[page=3,width=0.89\textwidth]{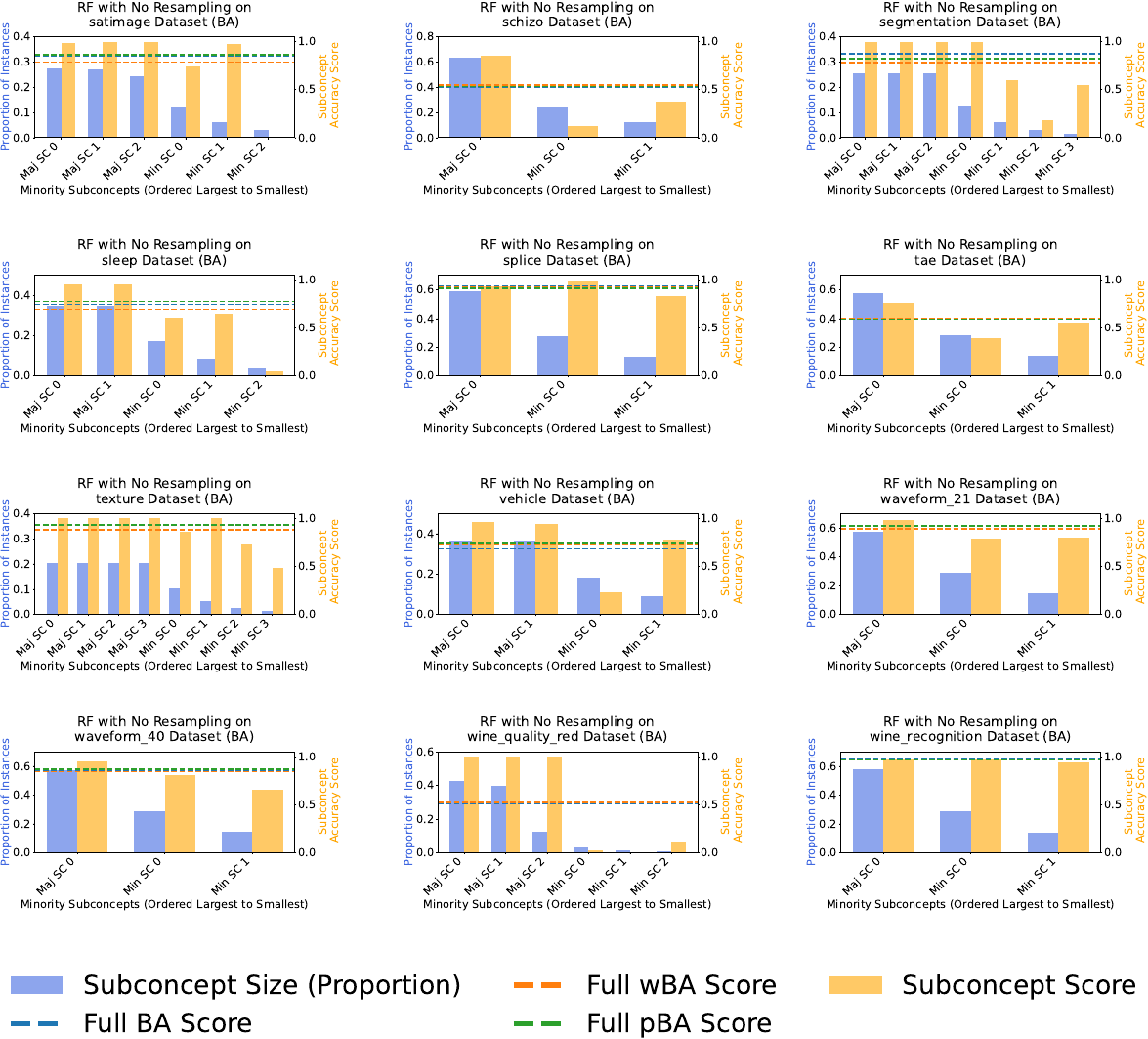}
    \caption{BA comparison plots on the 48 PMLB datasets (part 3 of 3). The interpretation of bars and lines is the same as in Figure \ref{fig:pmlb_ba_subconcept_plots}.}
    \label{fig:pmlb_ba_subconcept_plots_3}
\end{figure*}


\begin{figure*}[p]
    \centering
    \includegraphics[page=1,width=0.89\textwidth]{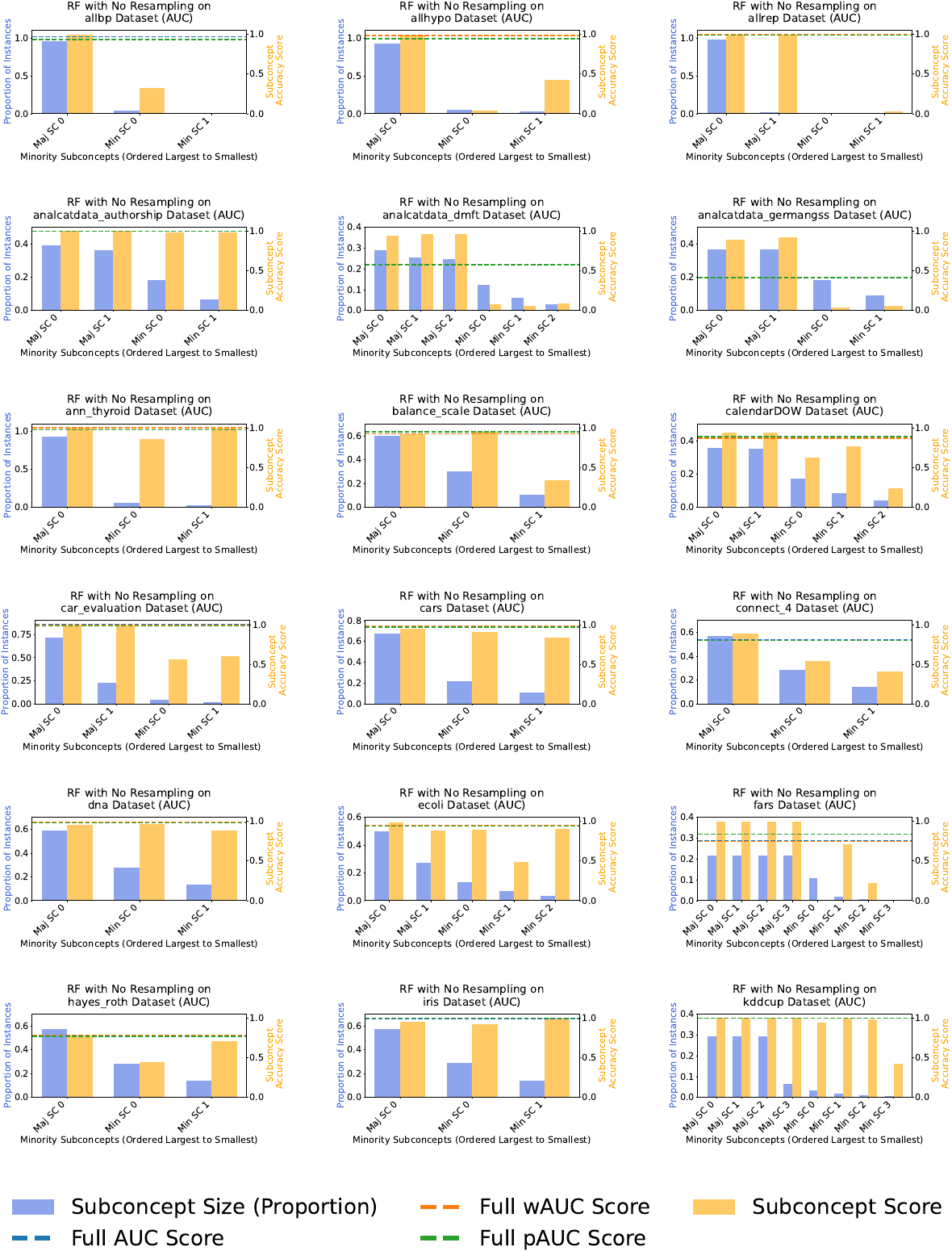}
    \caption{ROC-AUC comparison plots on the 48 PMLB datasets (part 1 of 3). The interpretation of bars and lines is the same as in Figure \ref{fig:pmlb_ba_subconcept_plots}.}
    \label{fig:pmlb_auc_subconcept_plots_1}
\end{figure*}

\begin{figure*}[p]
    \centering
    \includegraphics[page=2,width=0.89\textwidth]{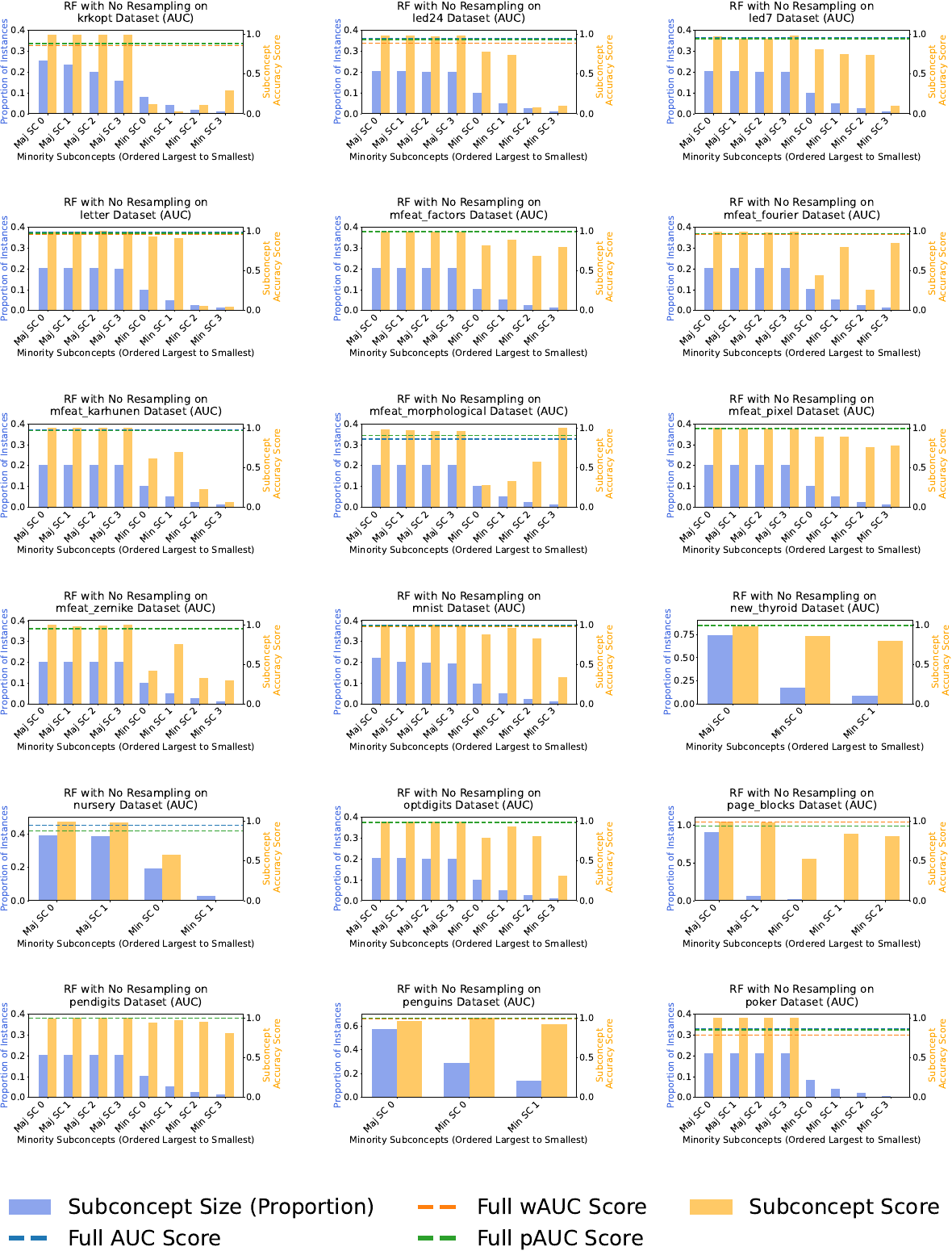}
    \caption{ROC-AUC comparison plots on the 48 PMLB datasets (part 2 of 3). The interpretation of bars and lines is the same as in Figure \ref{fig:pmlb_ba_subconcept_plots}.}
    \label{fig:pmlb_auc_subconcept_plots_2}
\end{figure*}

\begin{figure*}[p]
    \centering
    \includegraphics[page=3,width=0.89\textwidth]{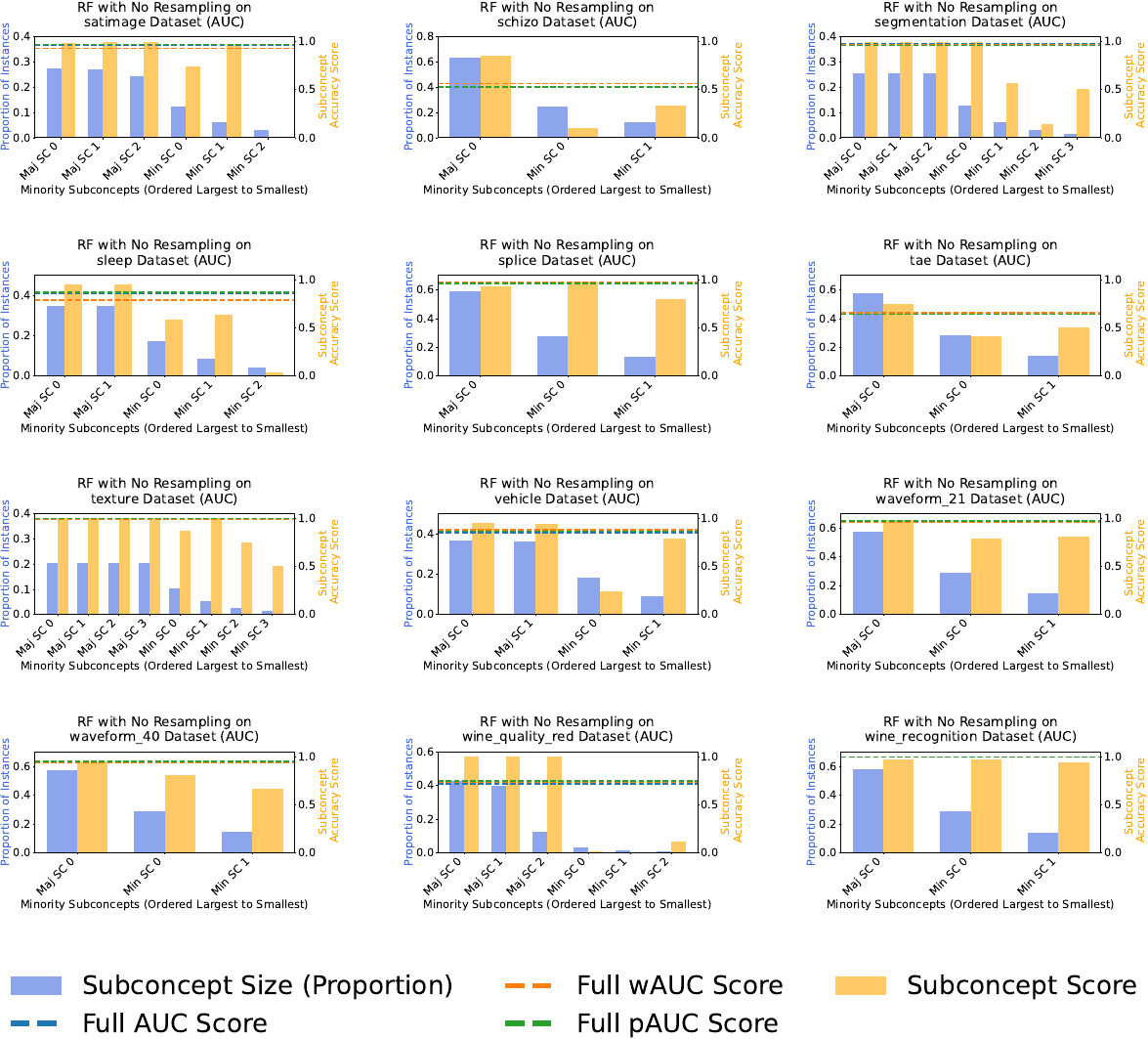}
    \caption{ROC-AUC comparison plots on the 48 PMLB datasets (part 3 of 3). The interpretation of bars and lines is the same as in Figure \ref{fig:pmlb_ba_subconcept_plots}.}
    \label{fig:pmlb_auc_subconcept_plots_3}
\end{figure*}

\begin{figure*}[p]
    \centering
    \includegraphics[page=1,width=0.89\textwidth]{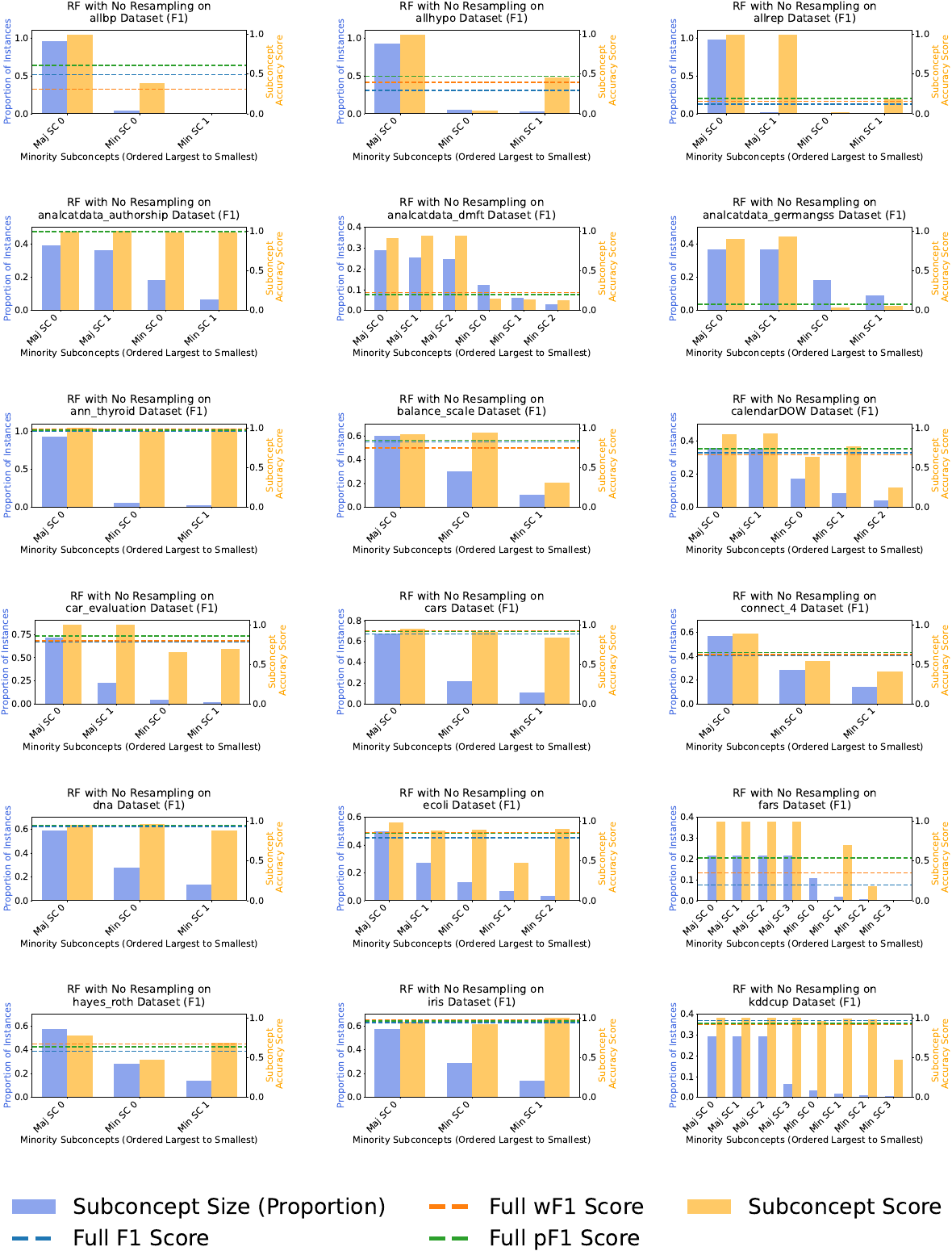}
    \caption{F1 comparison plots on the 48 PMLB datasets (part 1 of 3). The interpretation of bars and lines is the same as in Figure \ref{fig:pmlb_ba_subconcept_plots}.}
    \label{fig:pmlb_f1_subconcept_plots_1}
\end{figure*}

\begin{figure*}[p]
    \centering
    \includegraphics[page=2,width=0.89\textwidth]{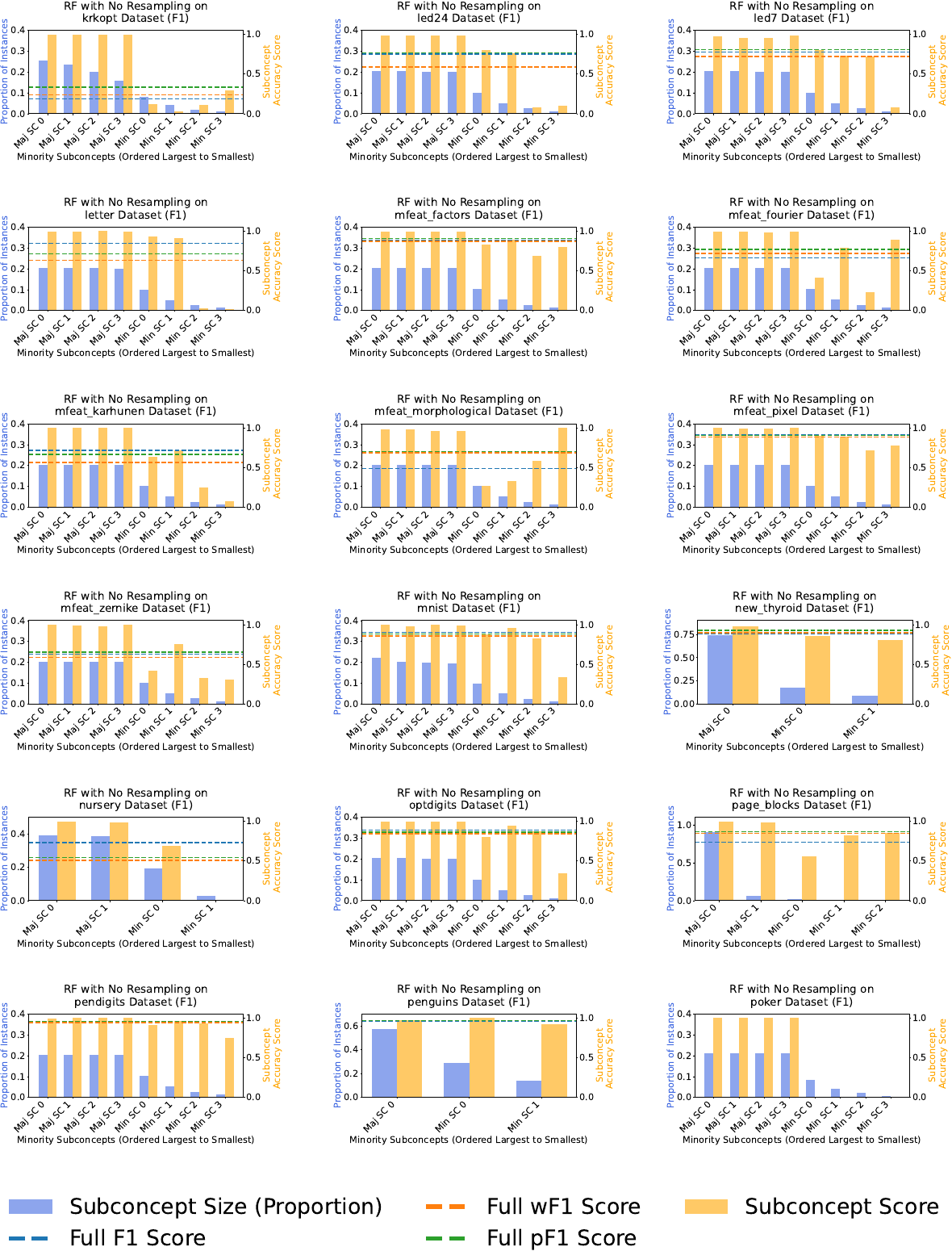}
    \caption{F1 comparison plots on the 48 PMLB datasets (part 2 of 3). The interpretation of bars and lines is the same as in Figure \ref{fig:pmlb_ba_subconcept_plots}.}
    \label{fig:pmlb_f1_subconcept_plots_2}
\end{figure*}

\begin{figure*}[p]
    \centering
    \includegraphics[page=3,width=0.89\textwidth]{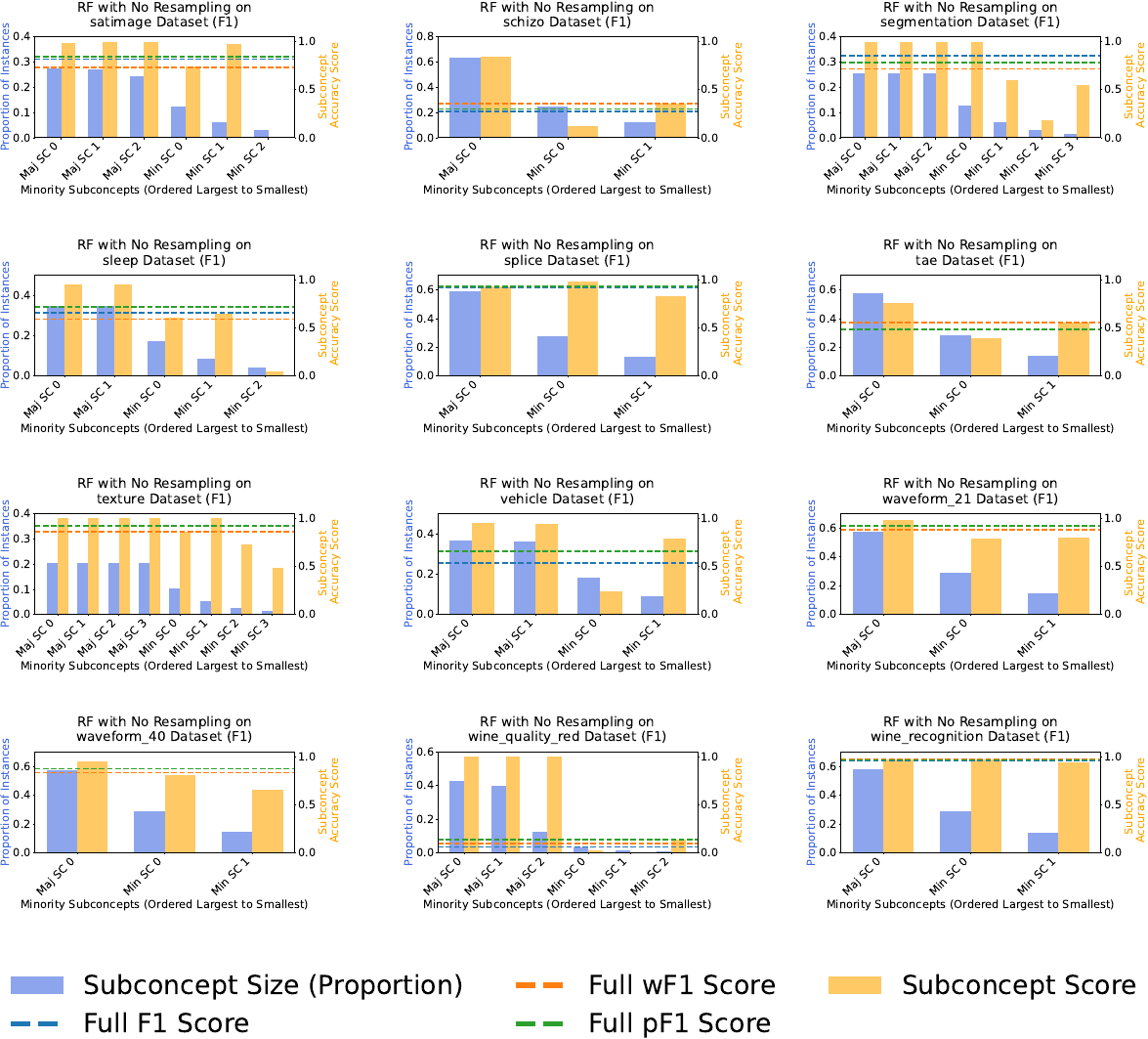}
    \caption{F1 comparison plots on the 48 PMLB datasets (part 3 of 3). The interpretation of bars and lines is the same as in Figure \ref{fig:pmlb_ba_subconcept_plots}.}
    \label{fig:pmlb_f1_subconcept_plots_3}
\end{figure*}


\begin{figure*}[p]
    \centering
    \setlength{\tabcolsep}{2pt}
    \begin{tabular}{ccc}
        \includegraphics[width=0.32\textwidth]{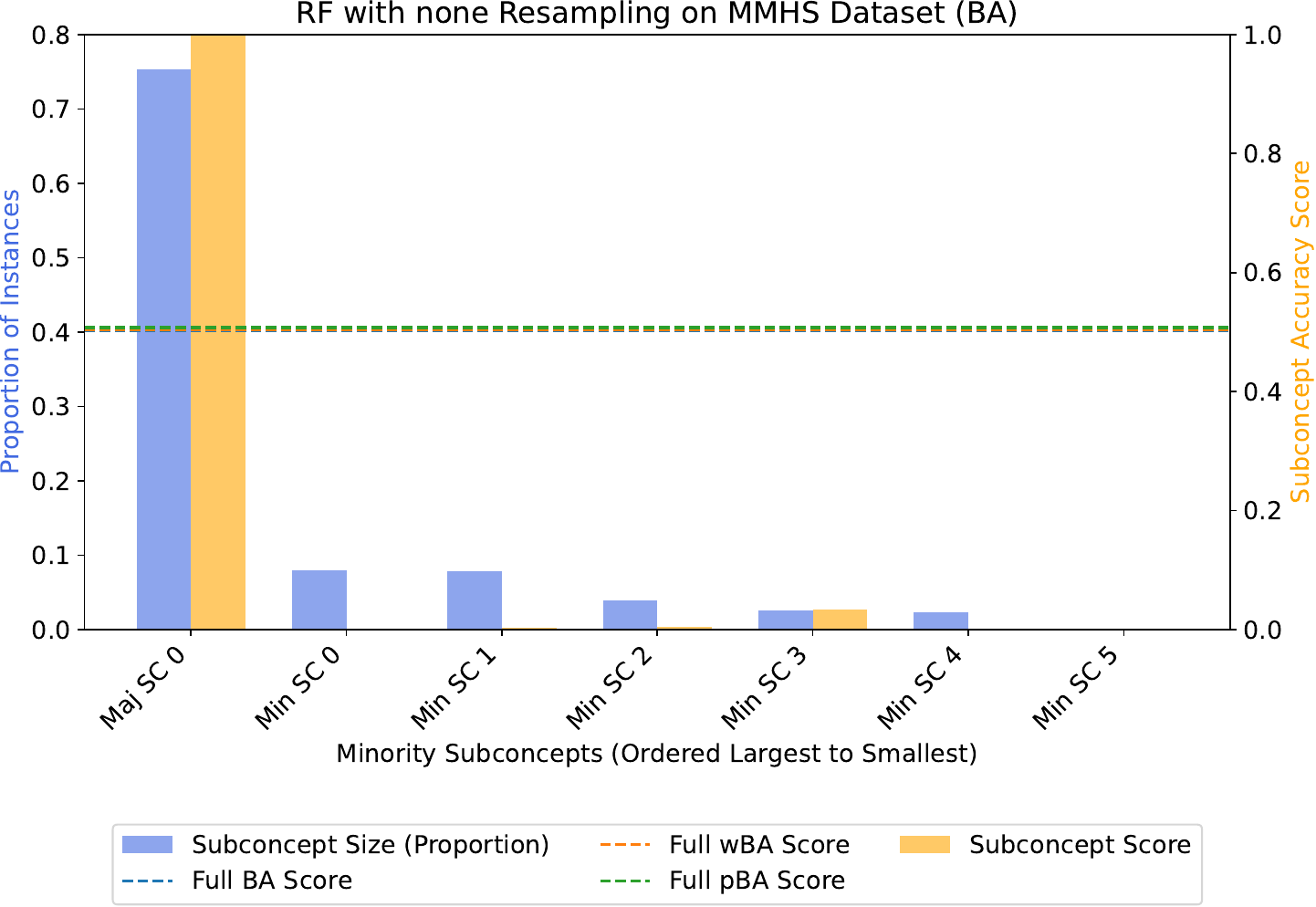} &
        \includegraphics[width=0.32\textwidth]{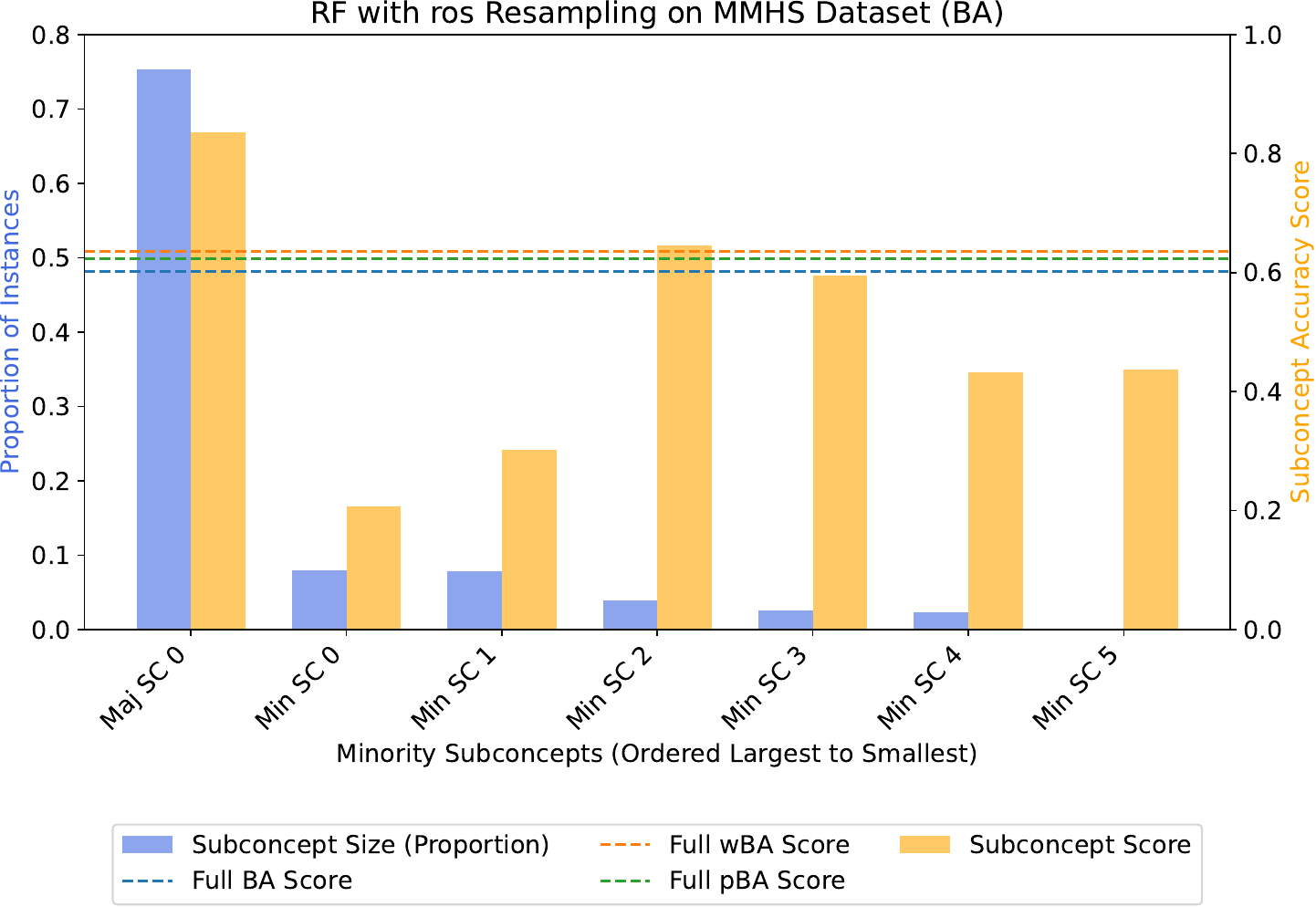} &
        \includegraphics[width=0.32\textwidth]{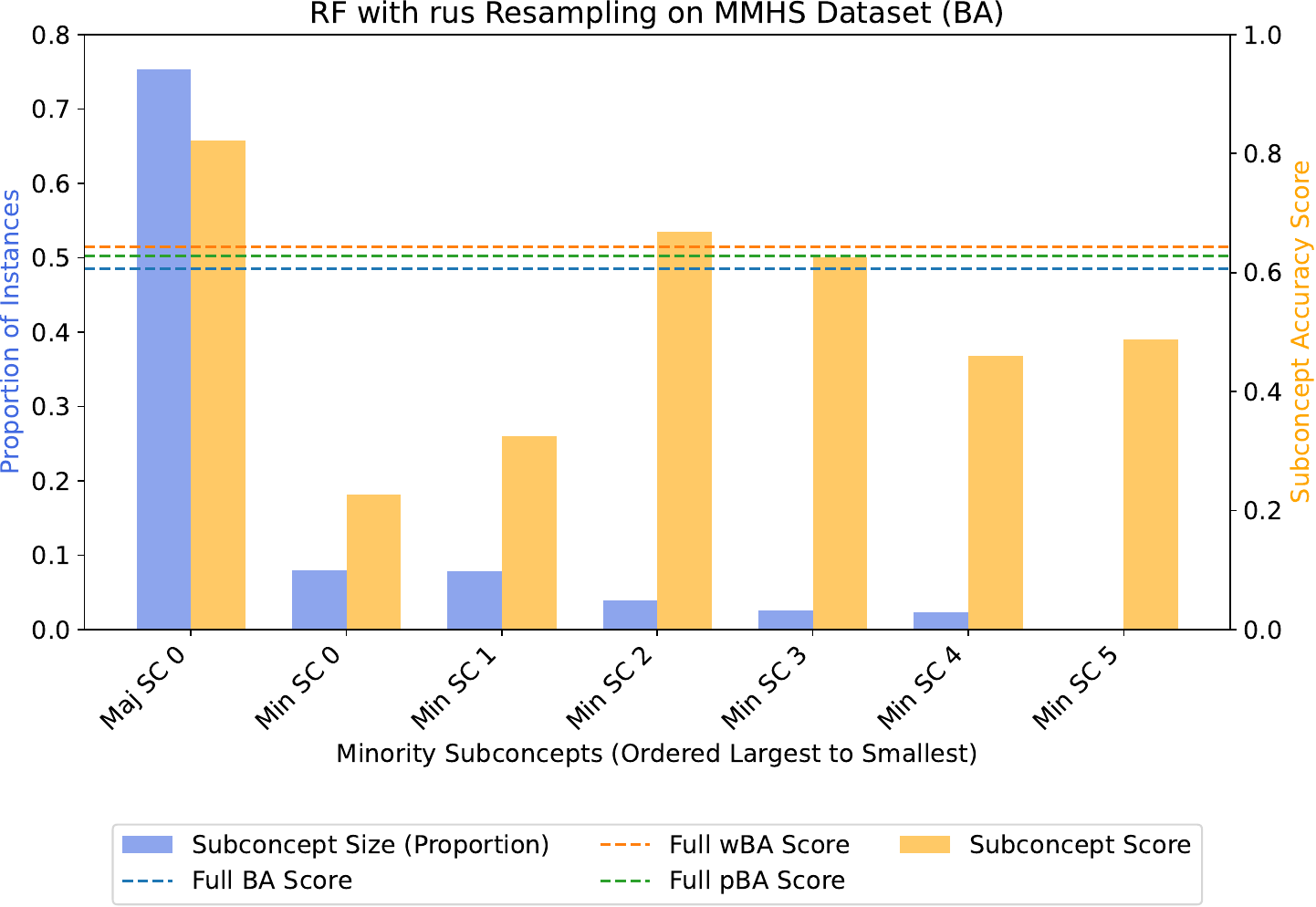} \\
        \includegraphics[width=0.32\textwidth]{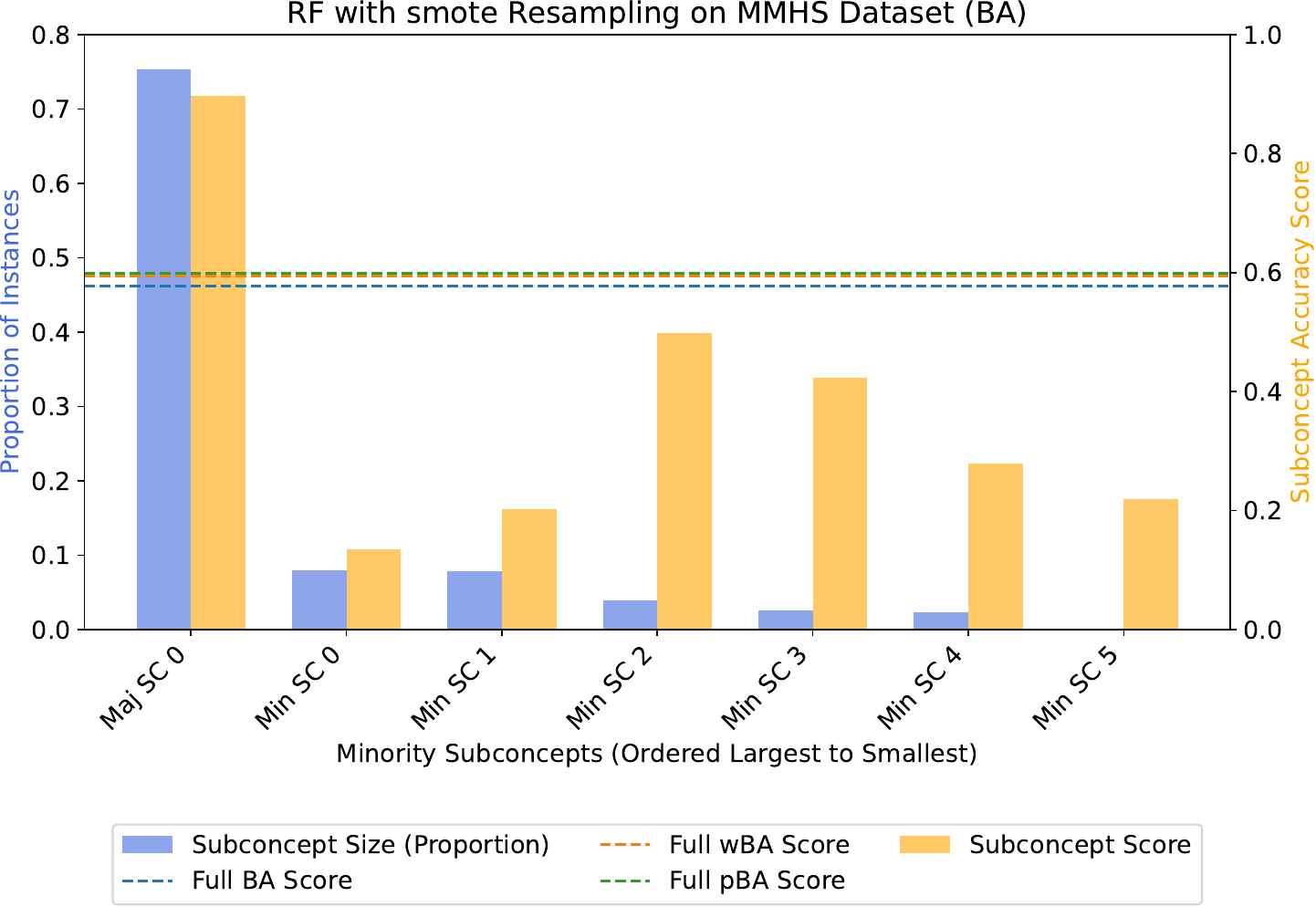} &
        \includegraphics[width=0.32\textwidth]{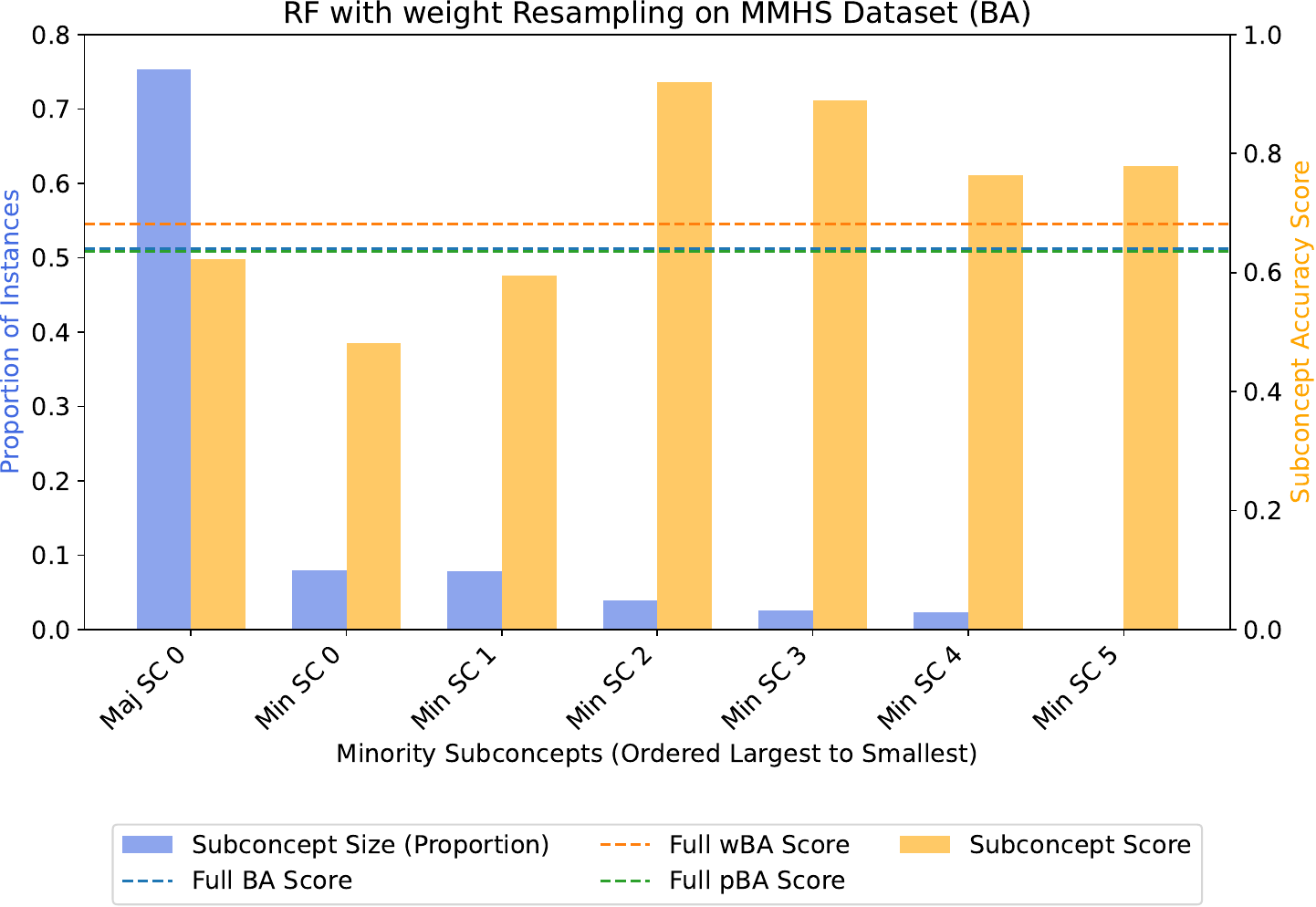} &
        \\
    \end{tabular}
    \caption{Balanced-accuracy comparison plots on MMHS150K under the five RF imbalance-correction settings. The interpretation of bars and lines is the same as in Figure \ref{fig:pmlb_ba_subconcept_plots}.}
    \label{fig:mmhs_ba_subconcept_plots}
\end{figure*}

\begin{figure*}[p]
    \centering
    \setlength{\tabcolsep}{2pt}
    \begin{tabular}{ccc}
        \includegraphics[width=0.32\textwidth]{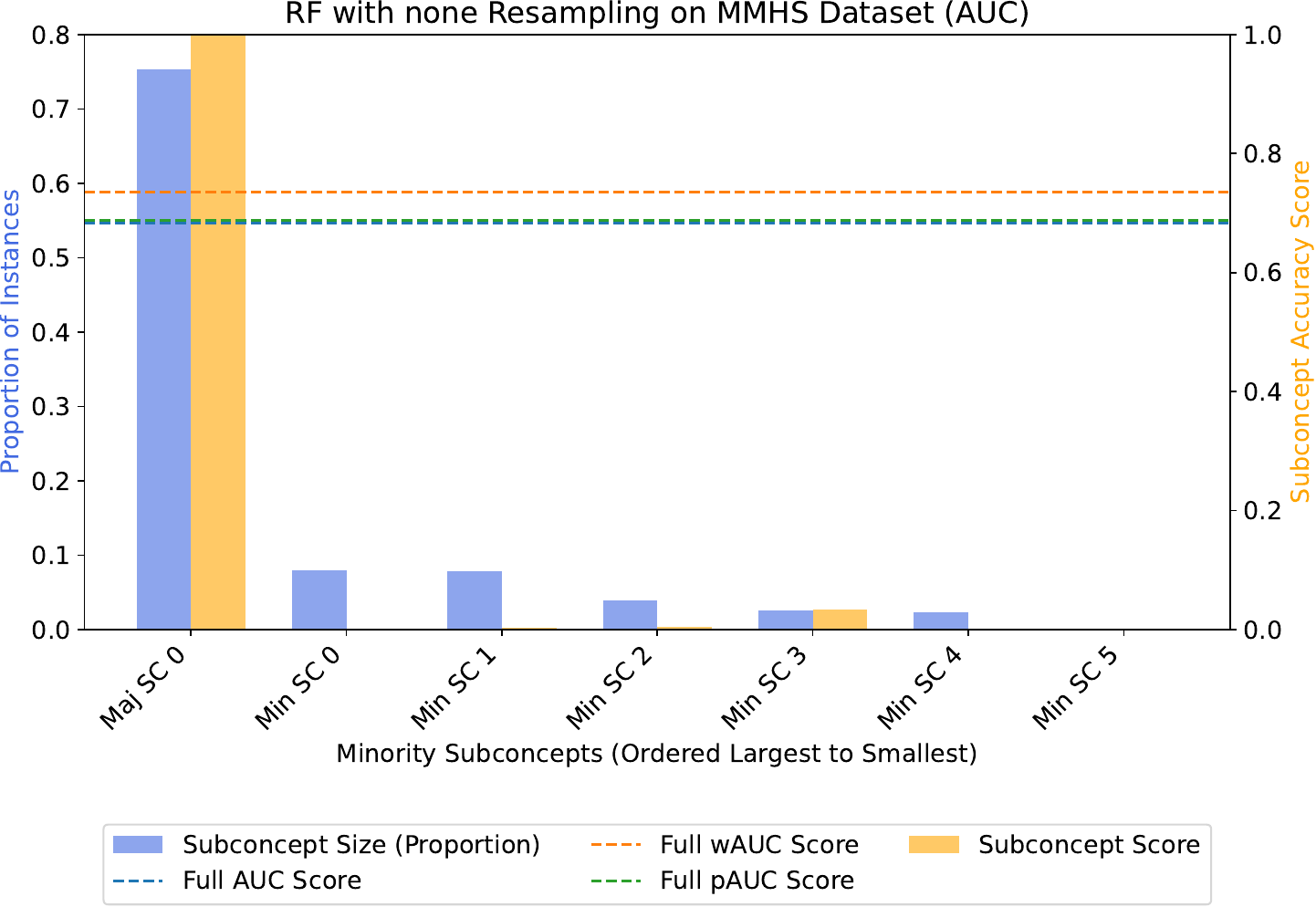} &
        \includegraphics[width=0.32\textwidth]{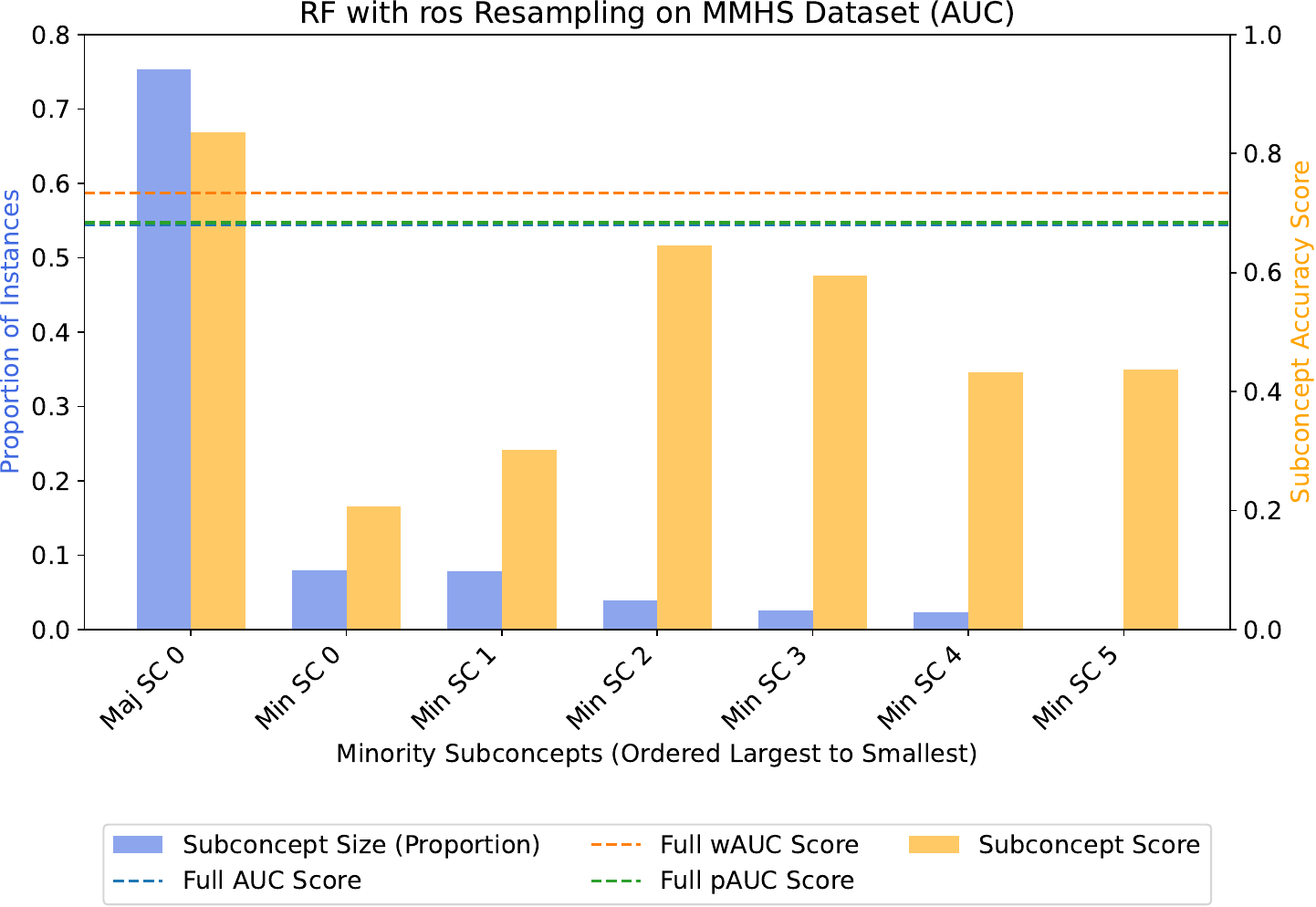} &
        \includegraphics[width=0.32\textwidth]{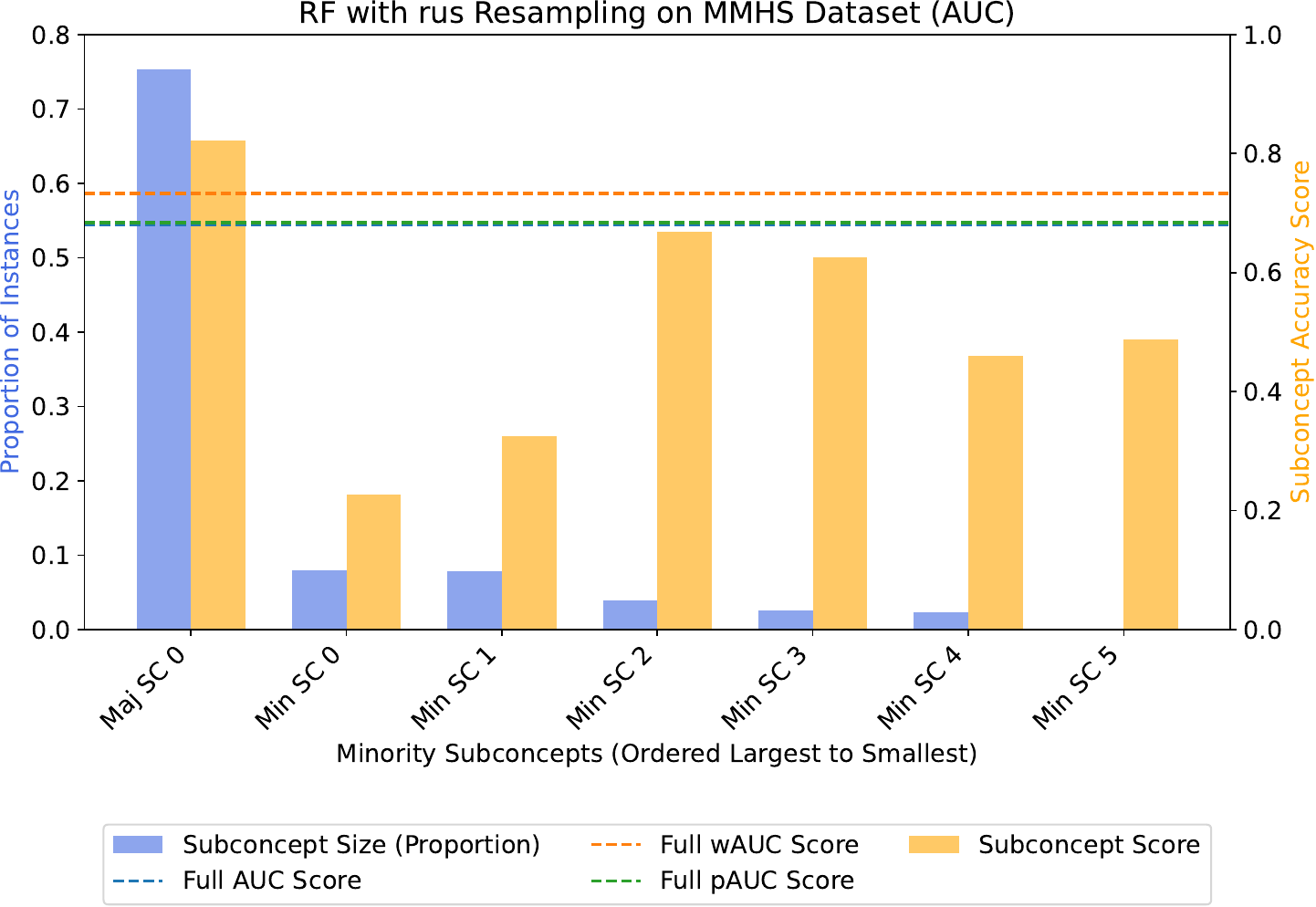} \\
        \includegraphics[width=0.32\textwidth]{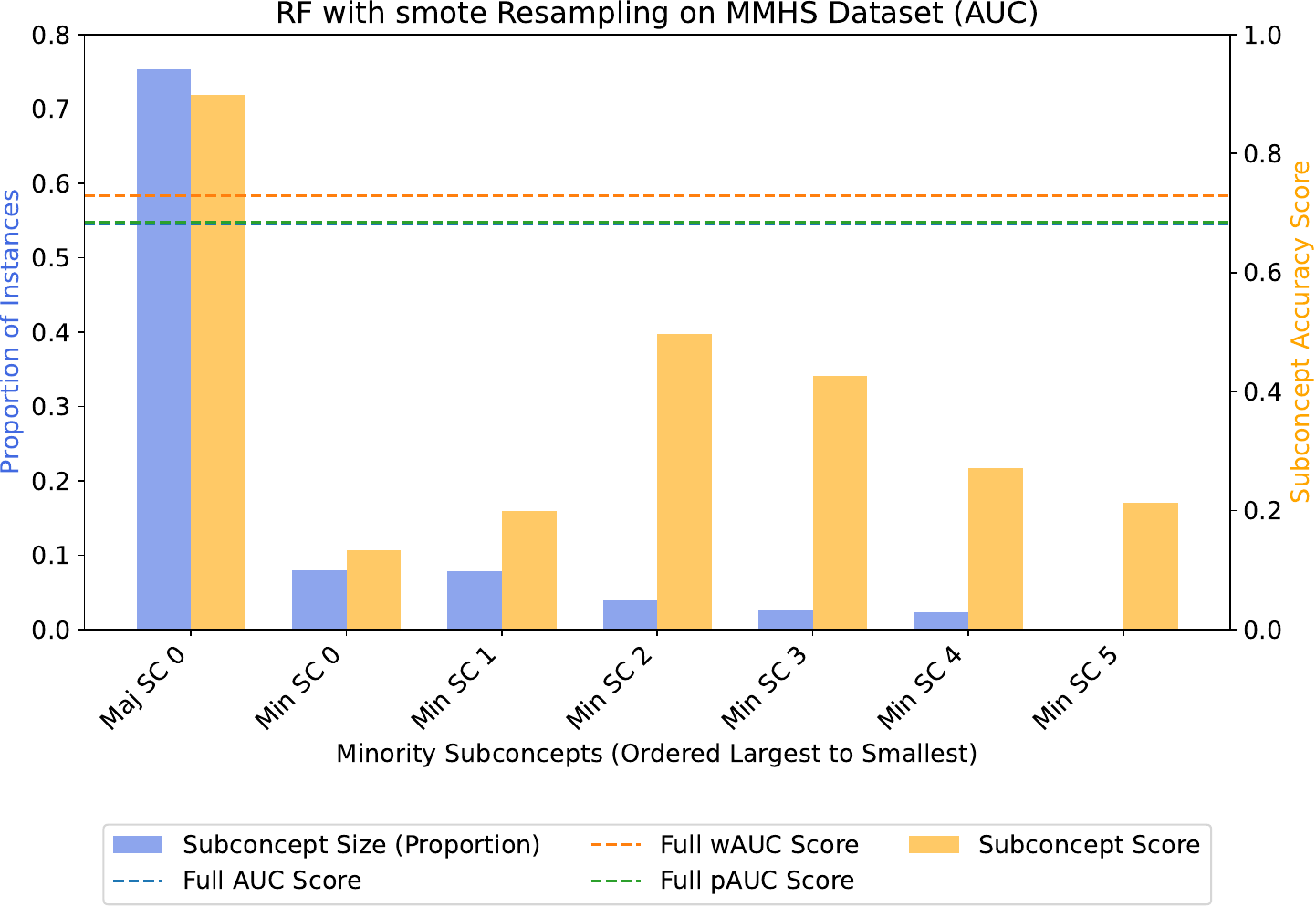} &
        \includegraphics[width=0.32\textwidth]{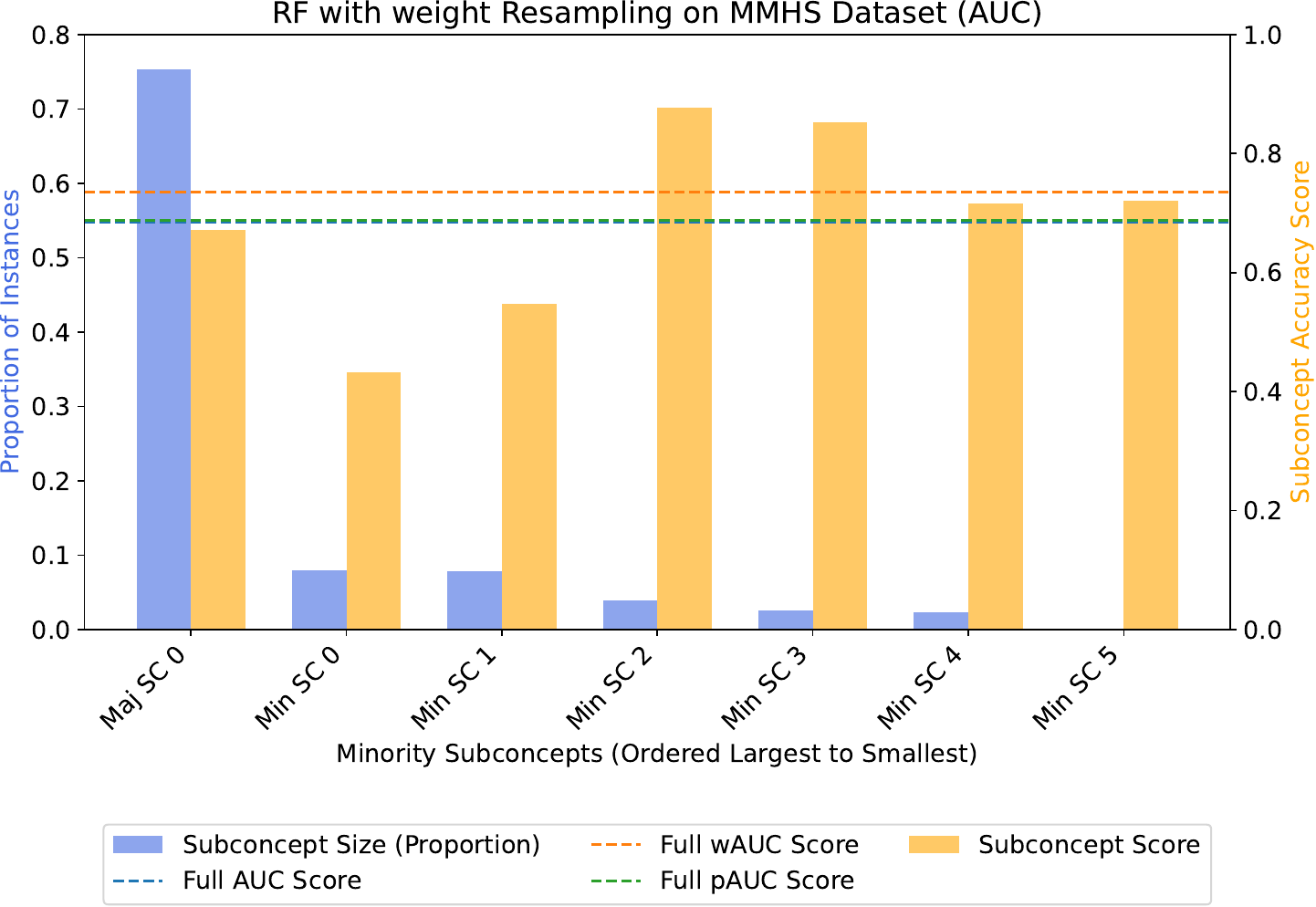} &
        \\
    \end{tabular}
    \caption{ROC-AUC comparison plots on MMHS150K under the five RF imbalance-correction settings. The interpretation of bars and lines is the same as in Figure \ref{fig:pmlb_ba_subconcept_plots}.}
    \label{fig:mmhs_auc_subconcept_plots}
\end{figure*}

\begin{figure*}[p]
    \centering
    \setlength{\tabcolsep}{2pt}
    \begin{tabular}{ccc}
        \includegraphics[width=0.32\textwidth]{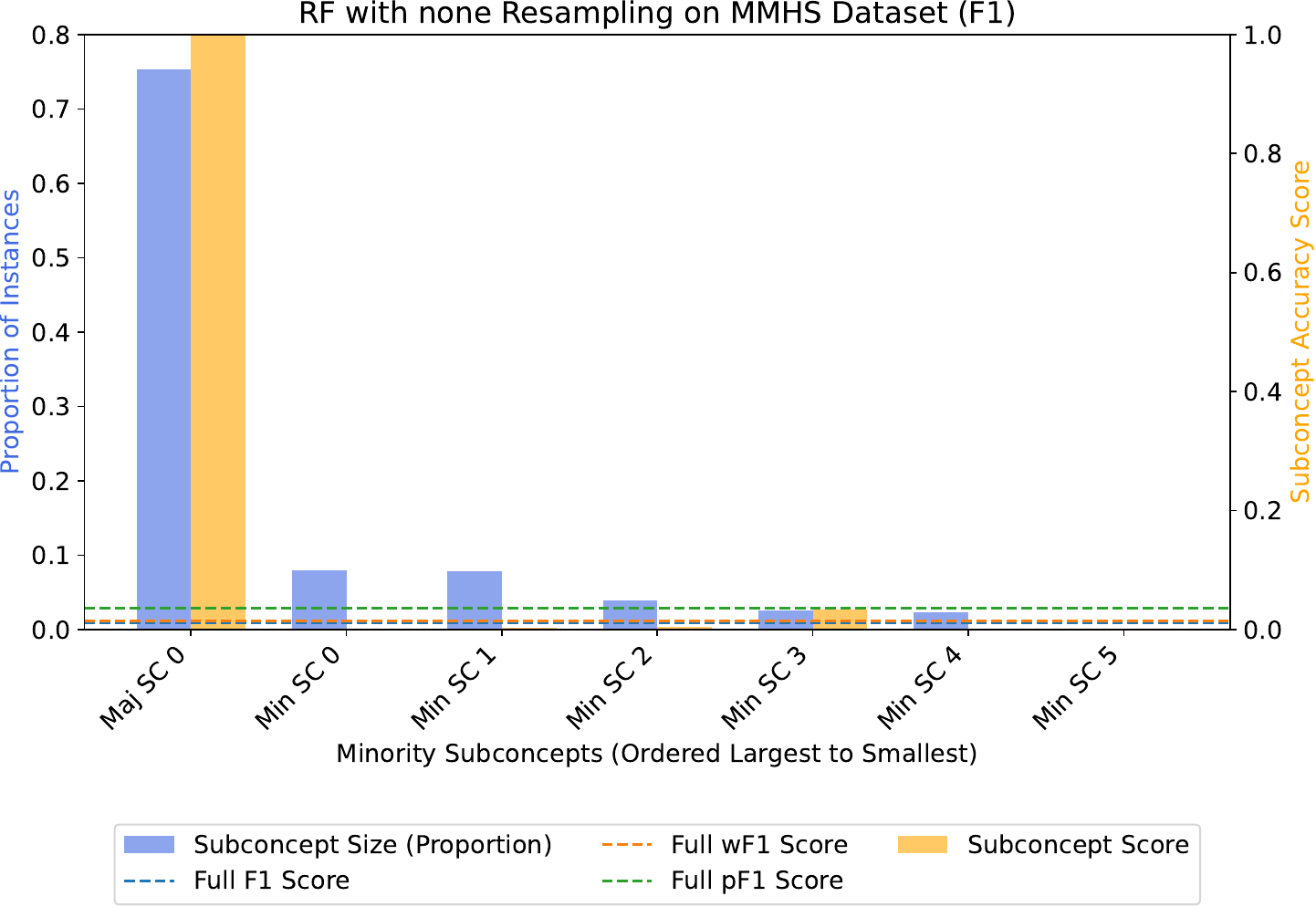} &
        \includegraphics[width=0.32\textwidth]{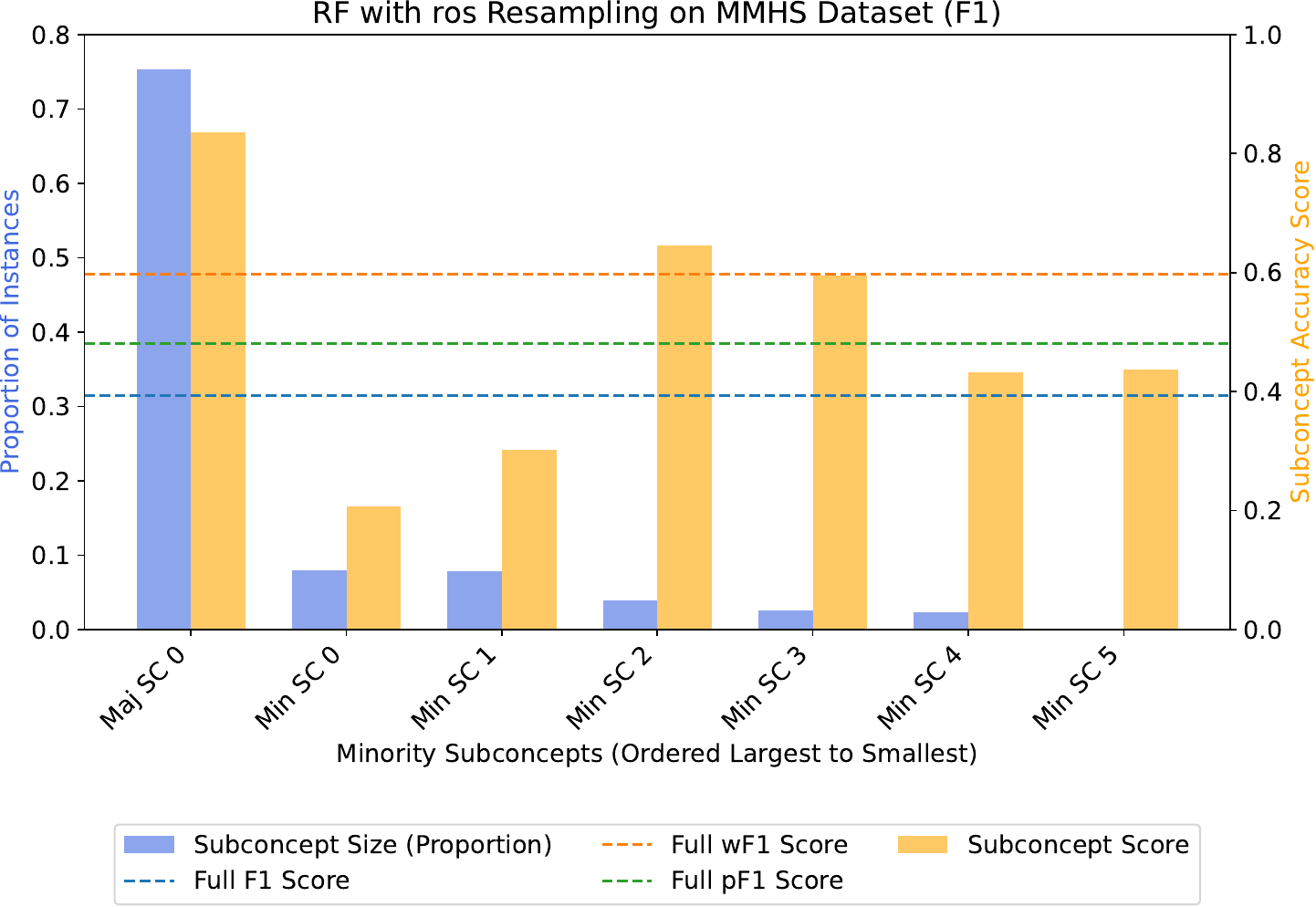} &
        \includegraphics[width=0.32\textwidth]{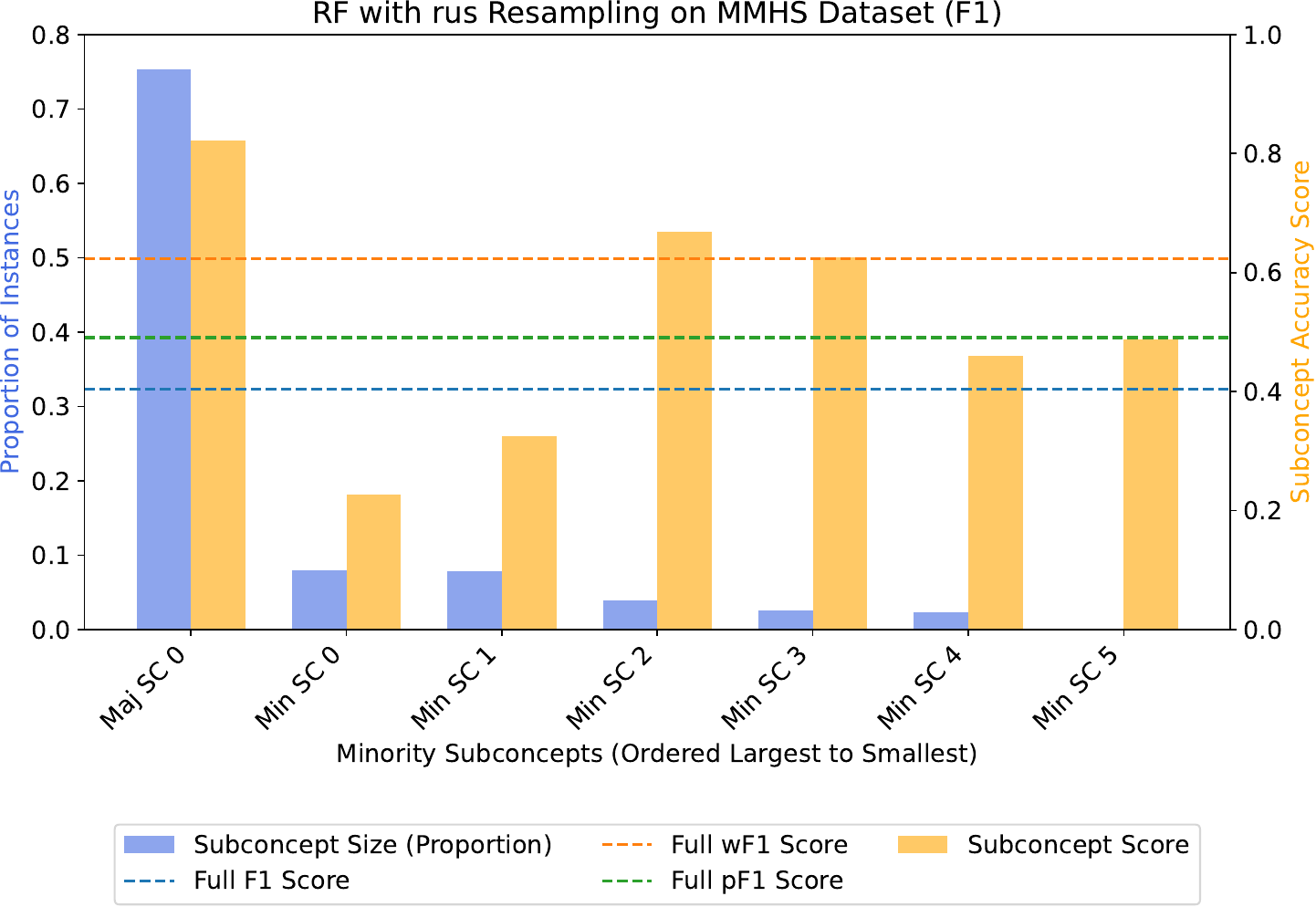} \\
        \includegraphics[width=0.32\textwidth]{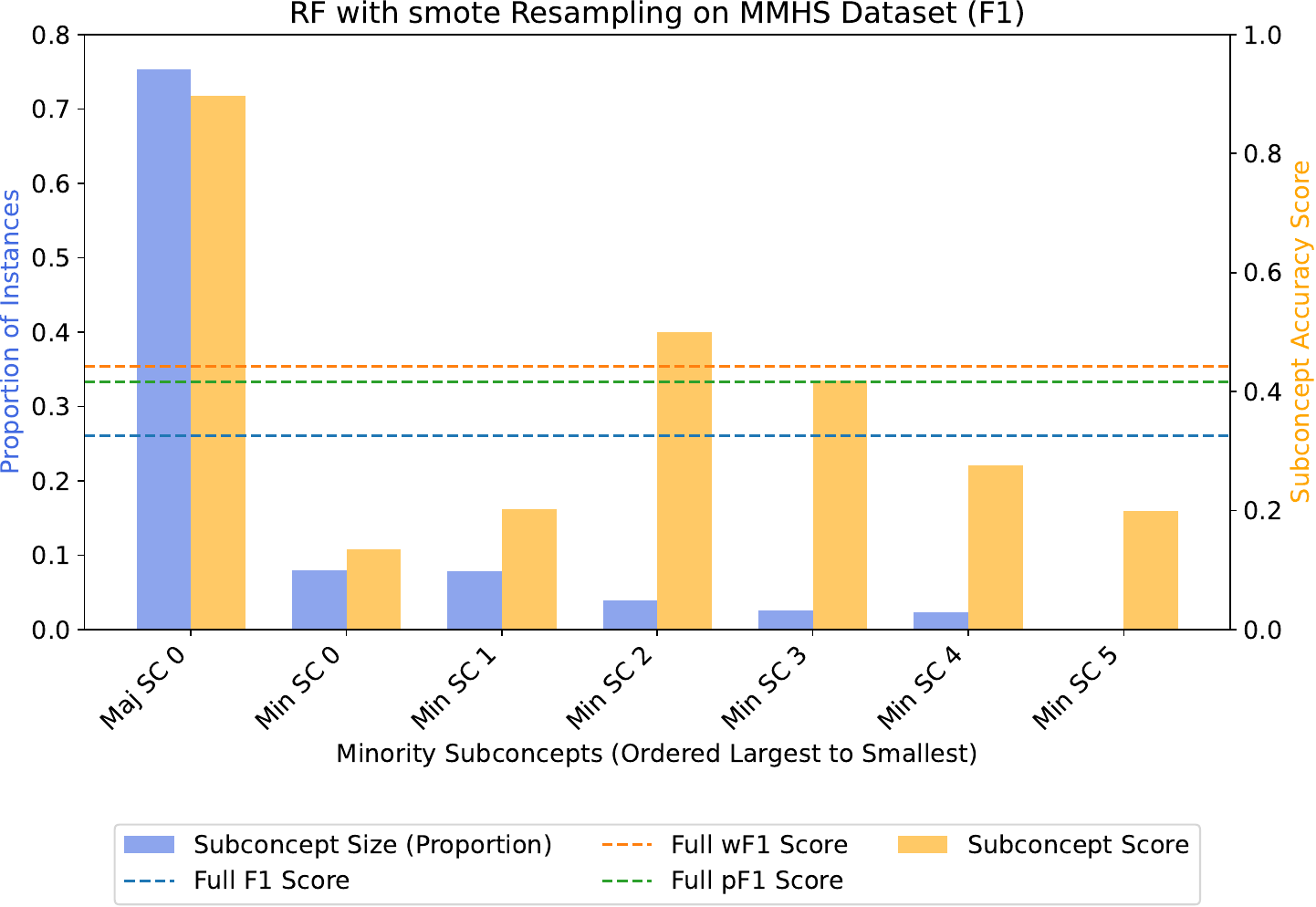} &
        \includegraphics[width=0.32\textwidth]{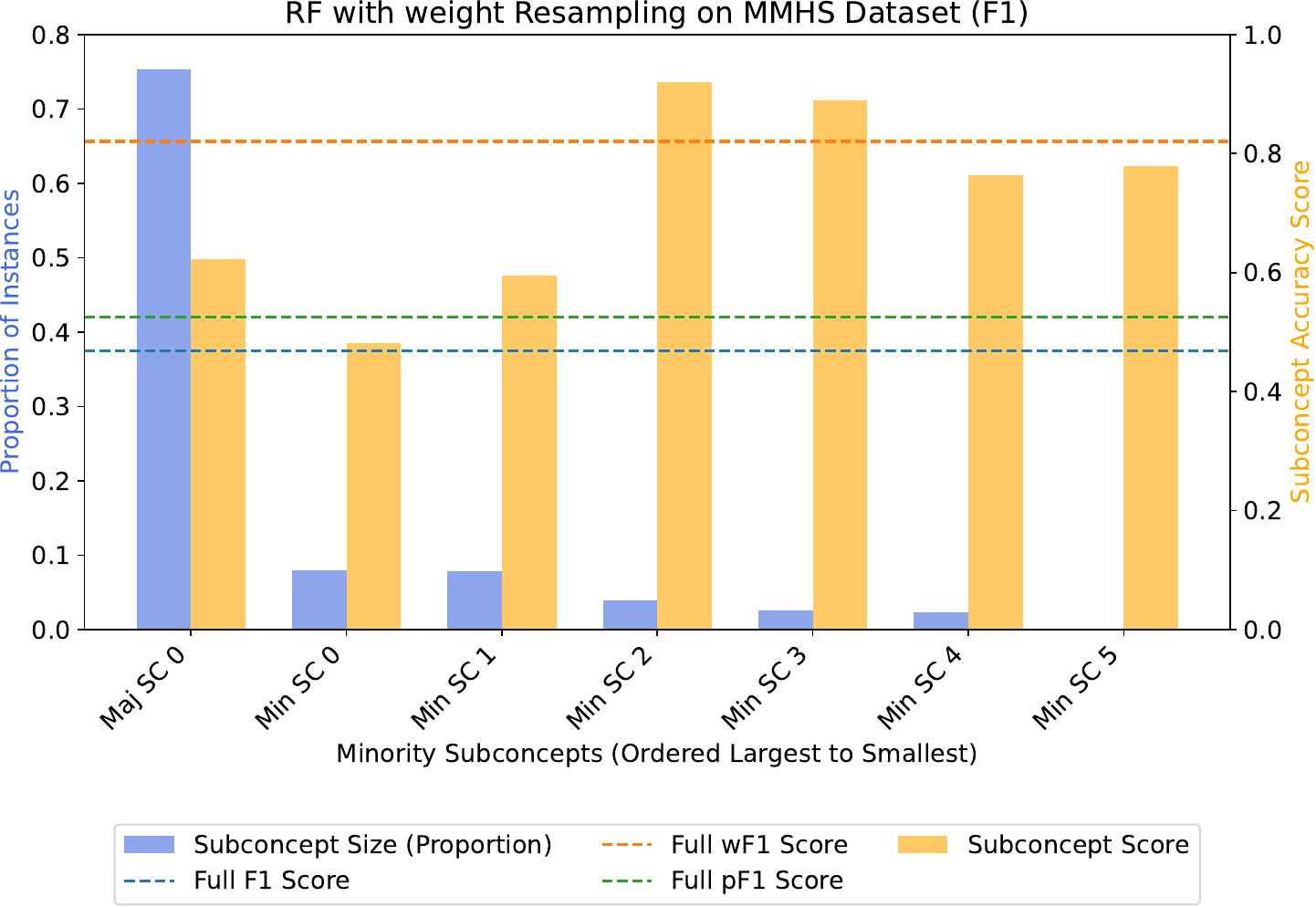} &
        \\
    \end{tabular}
    \caption{F1 comparison plots on MMHS150K under the five RF imbalance-correction settings. The interpretation of bars and lines is the same as in Figure \ref{fig:pmlb_ba_subconcept_plots}.}
    \label{fig:mmhs_f1_subconcept_plots}
\end{figure*}

\newcommand{\appendixmedicalplot}[1]{\includegraphics[width=\linewidth,trim=12 12 12 12,clip]{#1}}

\begin{figure*}[t]
    \centering
    \begin{minipage}{0.49\linewidth}
        \centering
        \appendixmedicalplot{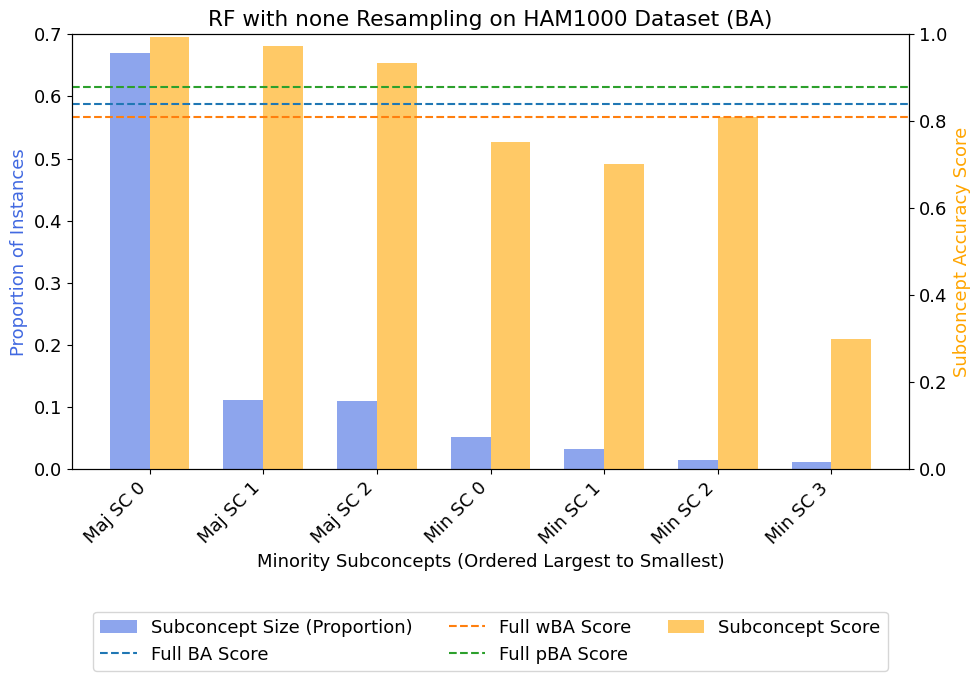}
    \end{minipage}
    \begin{minipage}{0.49\linewidth}
        \centering
        \appendixmedicalplot{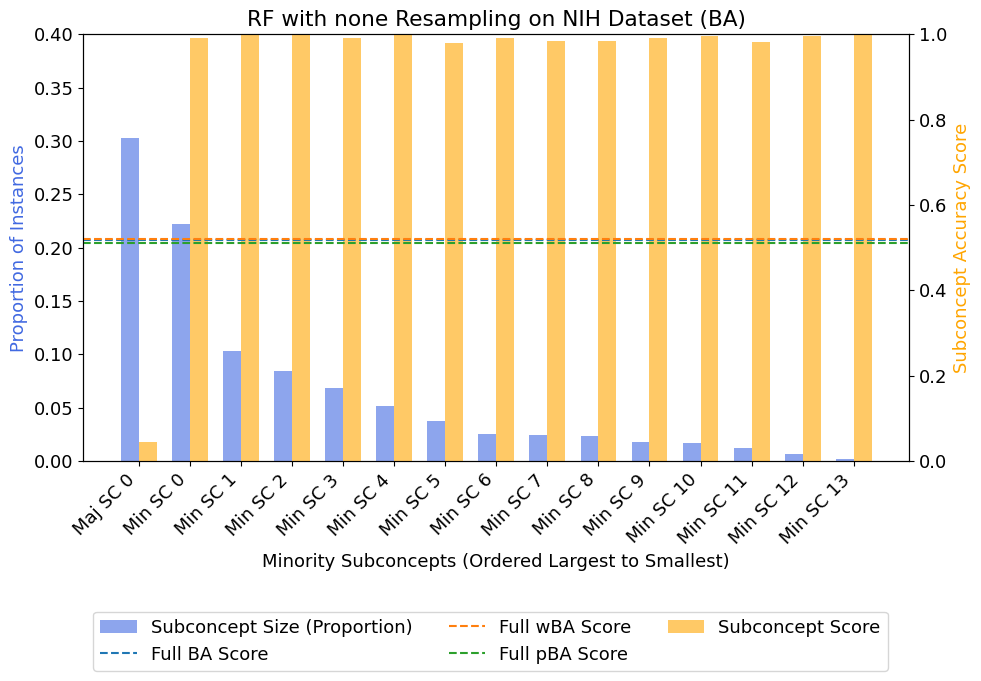}
    \end{minipage}

    \vspace{0.4em}

    \begin{minipage}{0.49\linewidth}
        \centering
        \appendixmedicalplot{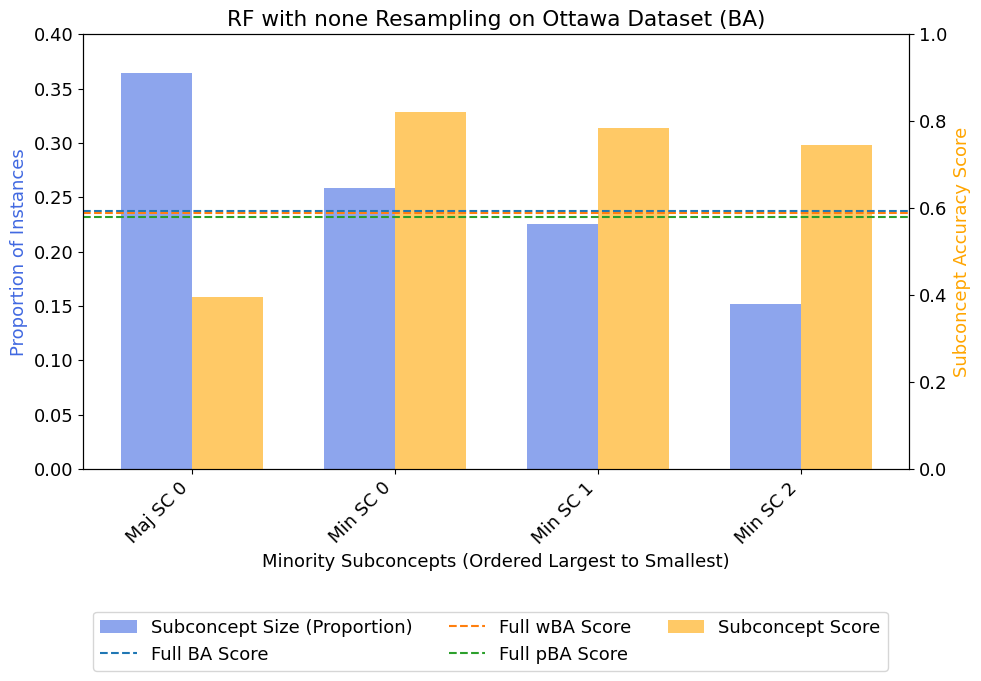}
    \end{minipage}
    \begin{minipage}{0.49\linewidth}
        \centering
        \appendixmedicalplot{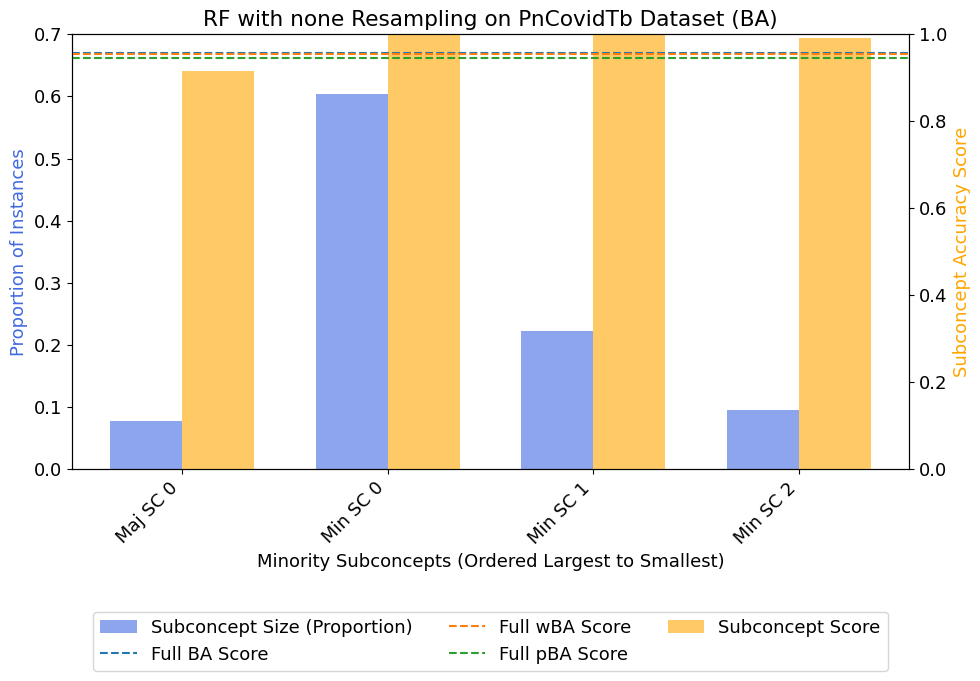}
    \end{minipage}
    \caption{Balanced-accuracy comparison plots on four medical datasets using embeddings followed by random forest classification. The solid blue bars show subconcept size, the yellow bars show subconcept accuracy, the dashed blue line shows the unweighted score, the green line shows the predicted-weight score, and the orange line shows the true-weight score.}
    \label{fig:medical_emb_rf_ba}
\end{figure*}

\begin{figure*}[t]
    \centering
    \begin{minipage}{0.49\linewidth}
        \centering
        \appendixmedicalplot{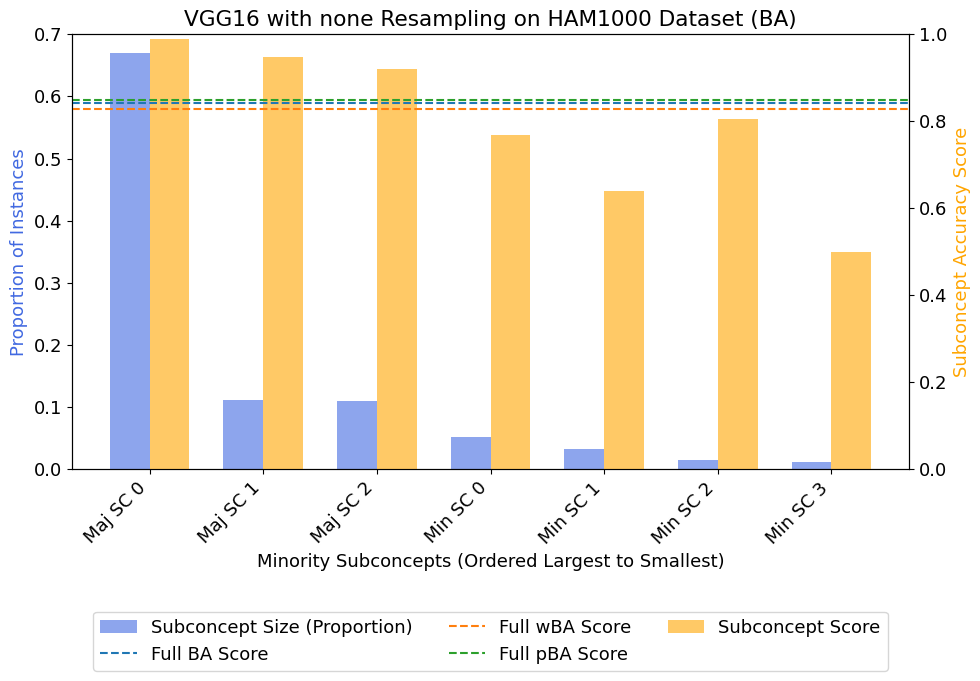}
    \end{minipage}
    \begin{minipage}{0.49\linewidth}
        \centering
        \appendixmedicalplot{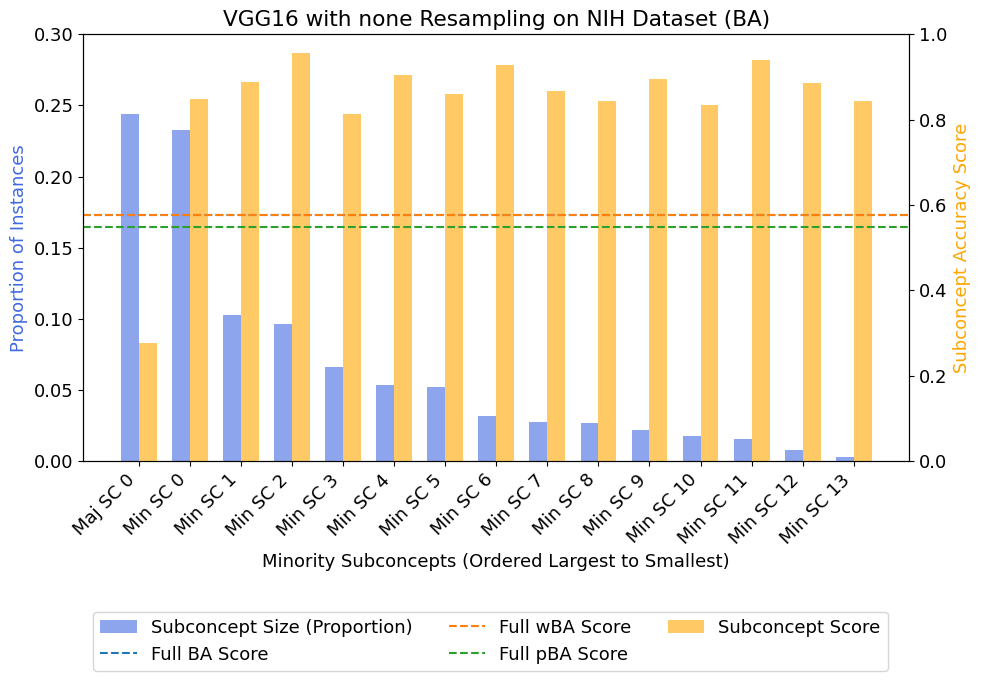}
    \end{minipage}

    \vspace{0.4em}

    \begin{minipage}{0.49\linewidth}
        \centering
        \appendixmedicalplot{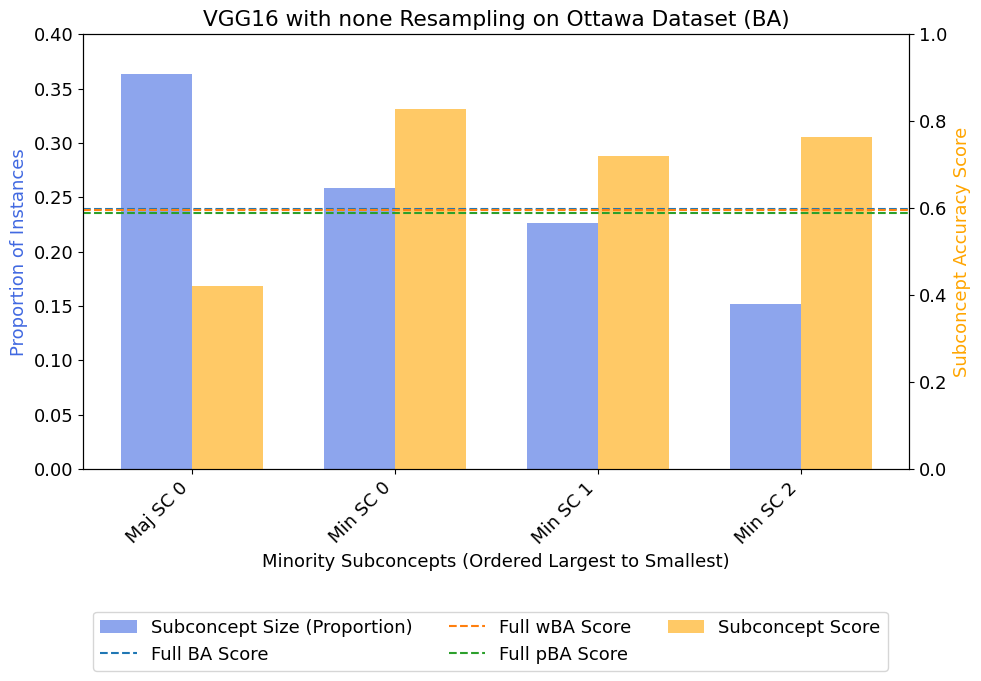}
    \end{minipage}
    \begin{minipage}{0.49\linewidth}
        \centering
        \appendixmedicalplot{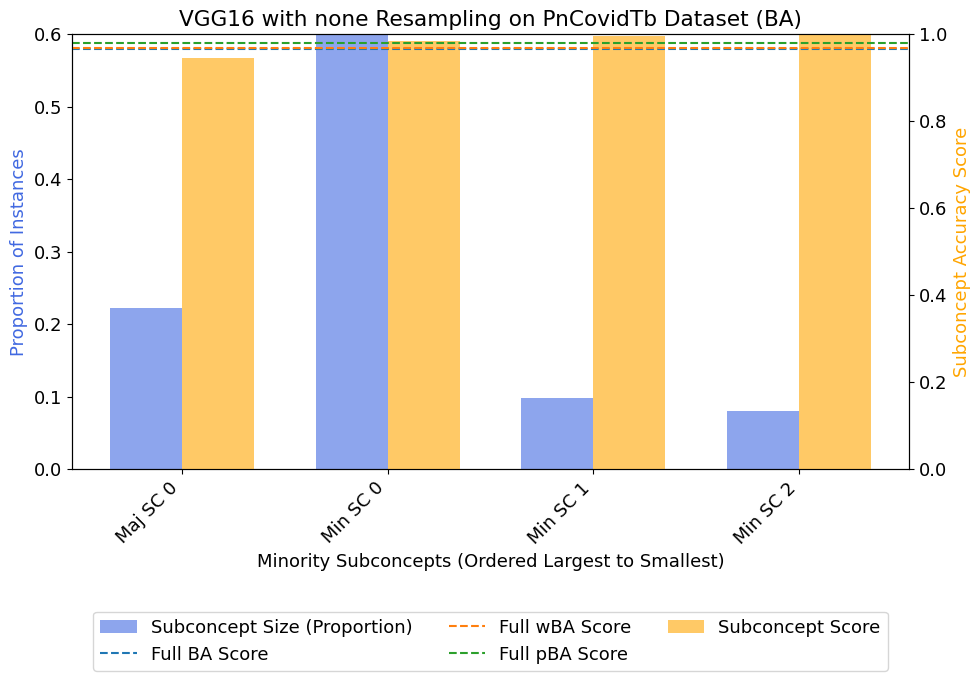}
    \end{minipage}
    \caption{Balanced-accuracy comparison plots on the same four medical datasets using direct VGG16 classification. The interpretation of colors and lines is the same as in Figure \ref{fig:medical_emb_rf_ba}.}
    \label{fig:medical_vgg_ba}
\end{figure*}

\begin{figure*}[t]
    \centering
    \begin{minipage}{0.49\linewidth}
        \centering
        \appendixmedicalplot{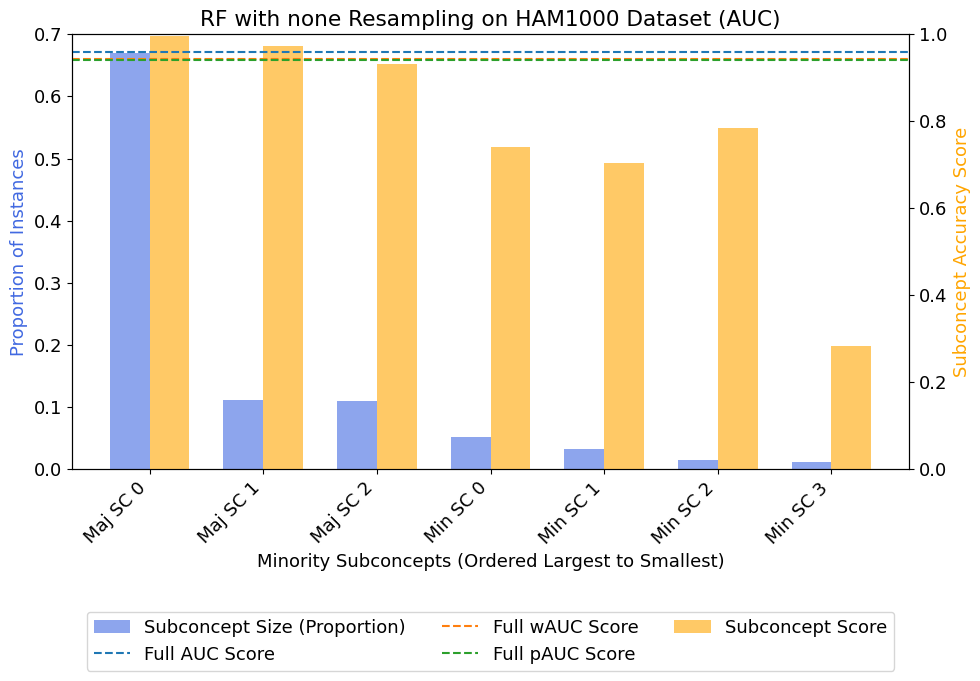}
    \end{minipage}
    \begin{minipage}{0.49\linewidth}
        \centering
        \appendixmedicalplot{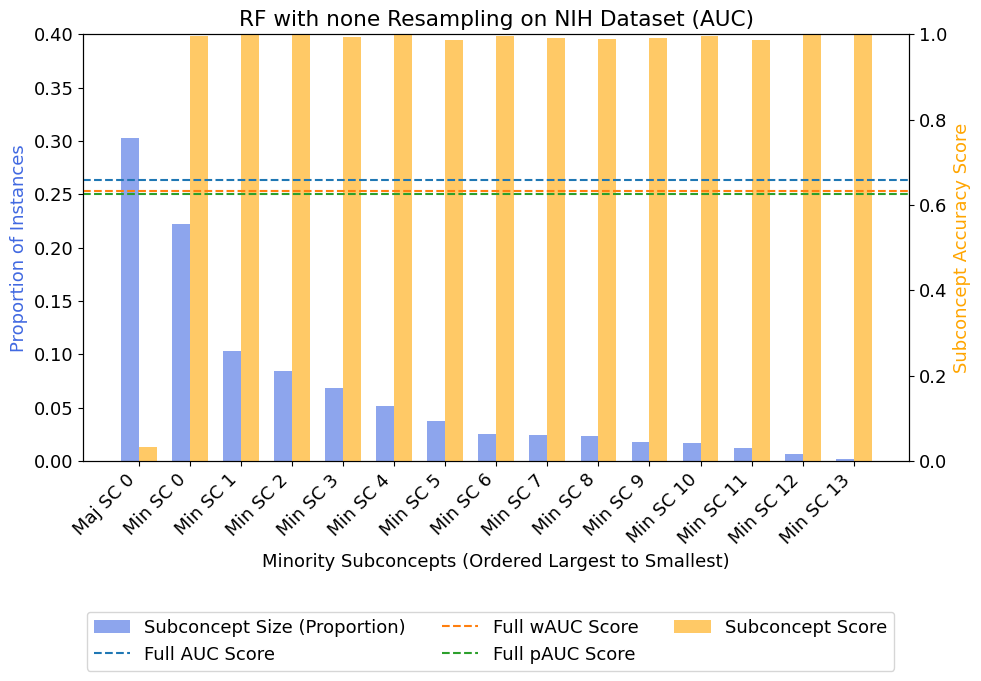}
    \end{minipage}

    \vspace{0.4em}

    \begin{minipage}{0.49\linewidth}
        \centering
        \appendixmedicalplot{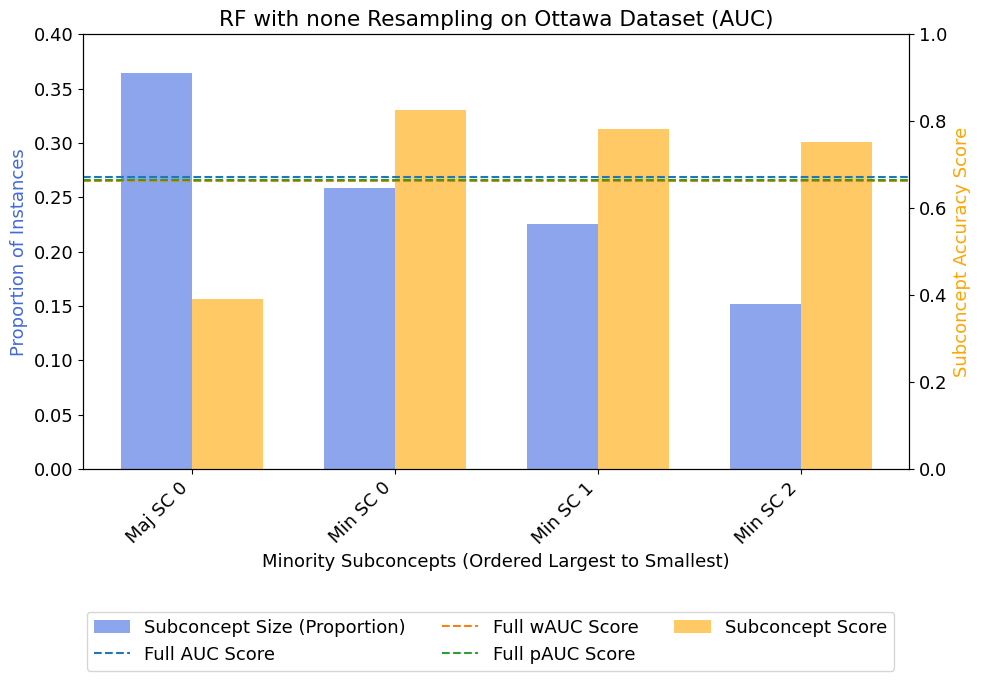}
    \end{minipage}
    \begin{minipage}{0.49\linewidth}
        \centering
        \appendixmedicalplot{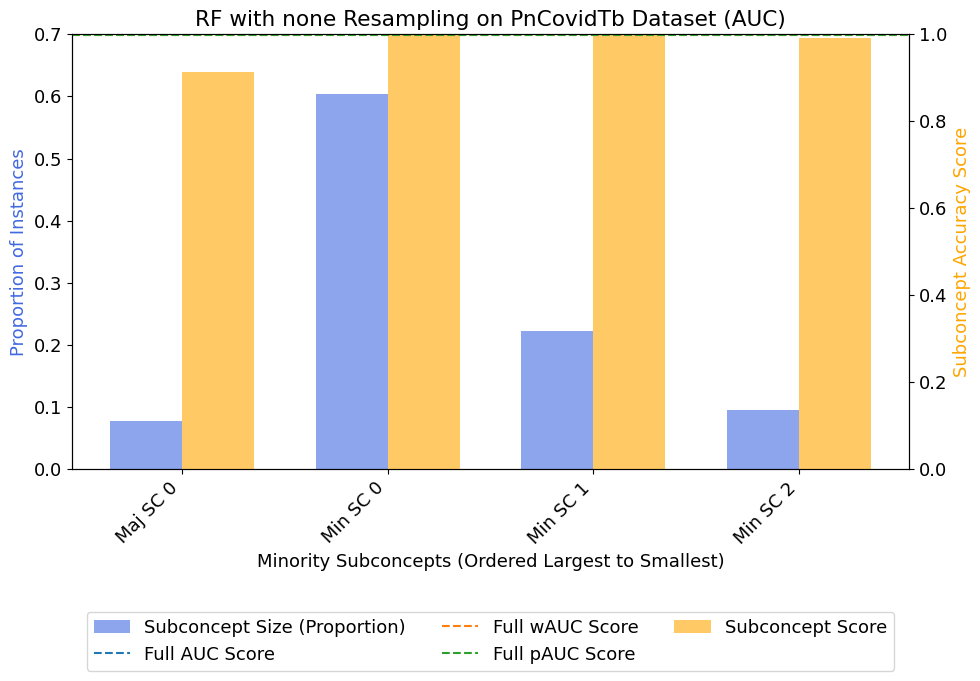}
    \end{minipage}
    \caption{ROC-AUC comparison plots on four medical datasets using embeddings followed by random forest classification. The interpretation of colors and lines is the same as in Figure \ref{fig:medical_emb_rf_ba}.}
    \label{fig:medical_emb_rf_auc}
\end{figure*}

\begin{figure*}[t]
    \centering
    \begin{minipage}{0.49\linewidth}
        \centering
        \appendixmedicalplot{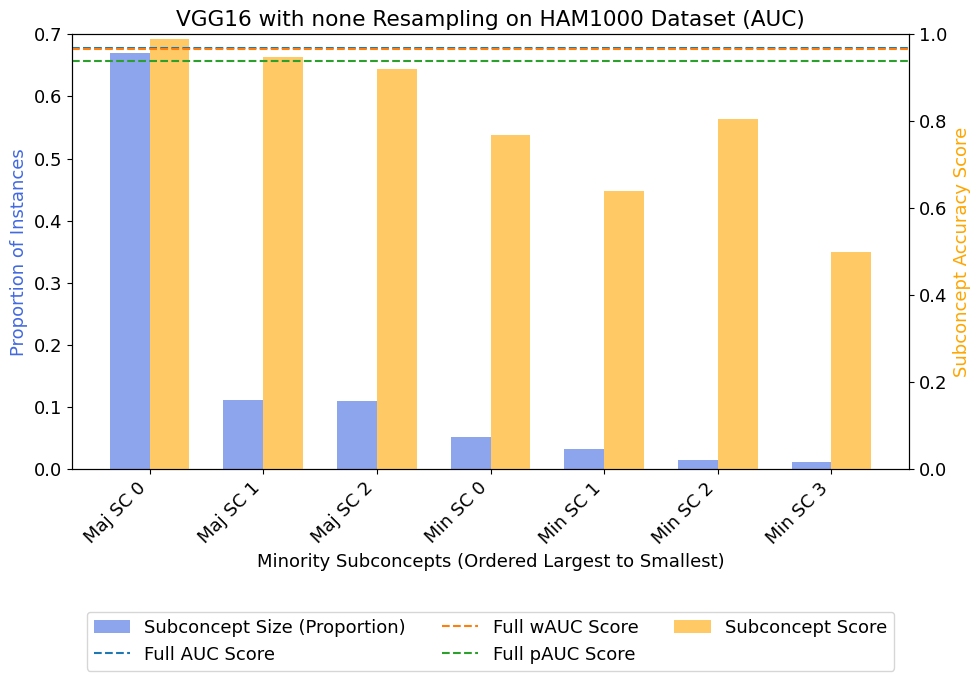}
    \end{minipage}
    \begin{minipage}{0.49\linewidth}
        \centering
        \appendixmedicalplot{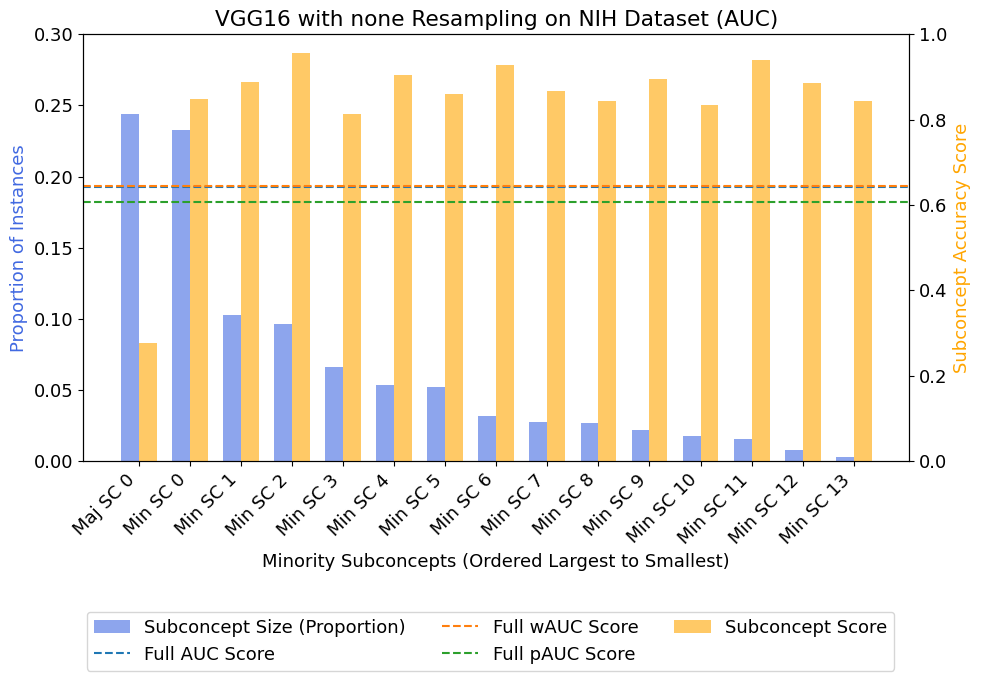}
    \end{minipage}

    \vspace{0.4em}

    \begin{minipage}{0.49\linewidth}
        \centering
        \appendixmedicalplot{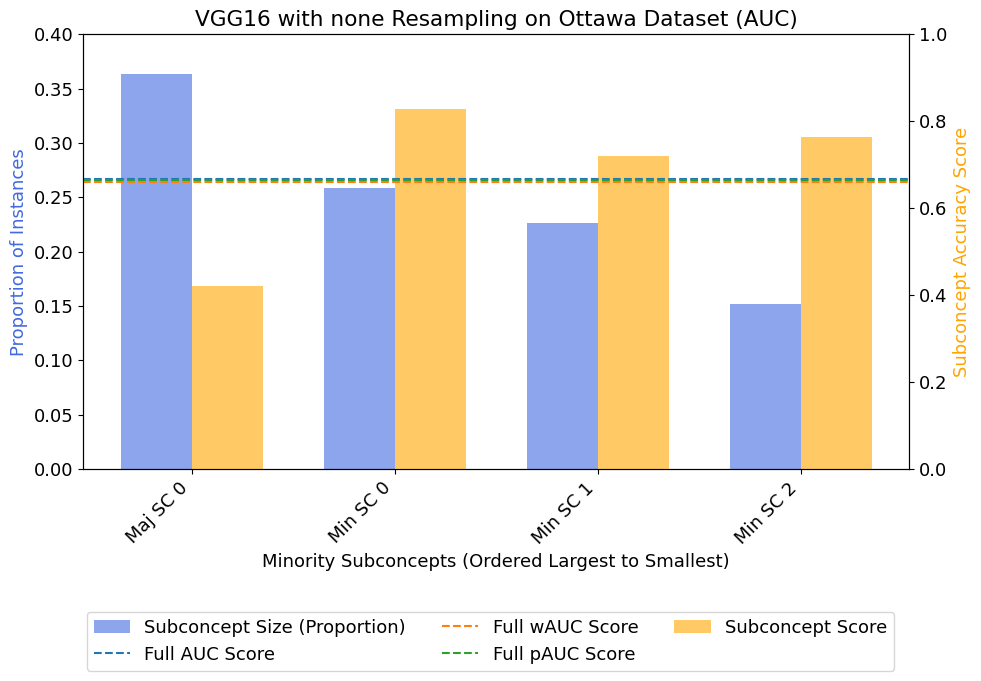}
    \end{minipage}
    \begin{minipage}{0.49\linewidth}
        \centering
        \appendixmedicalplot{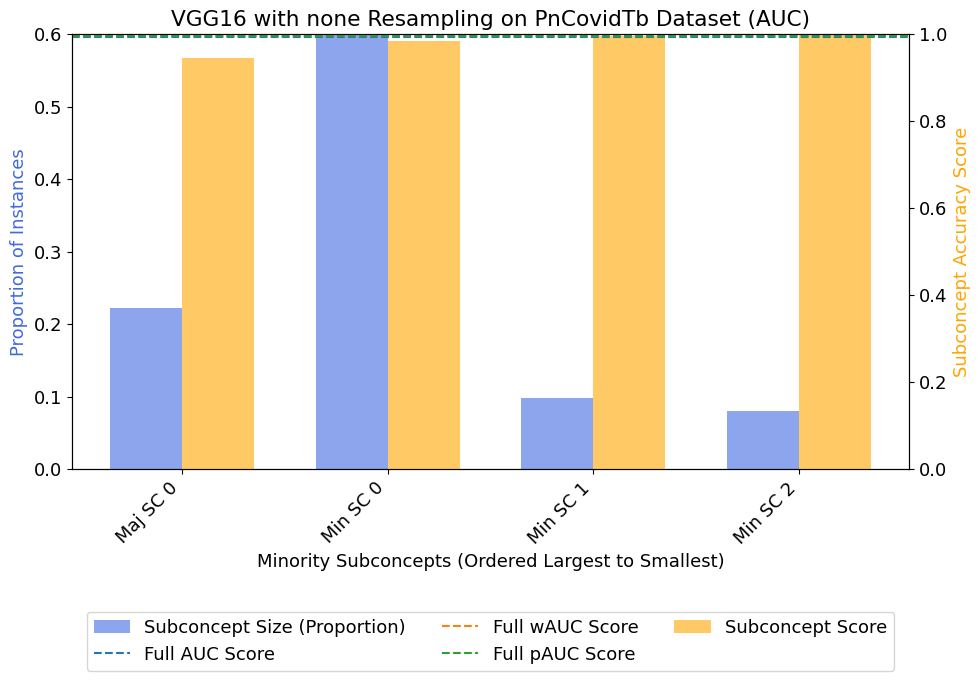}
    \end{minipage}
    \caption{ROC-AUC comparison plots on the same four medical datasets using direct VGG16 classification. The interpretation of colors and lines is the same as in Figure \ref{fig:medical_emb_rf_ba}.}
    \label{fig:medical_vgg_auc}
\end{figure*}

\begin{figure*}[t]
    \centering
    \begin{minipage}{0.49\linewidth}
        \centering
        \appendixmedicalplot{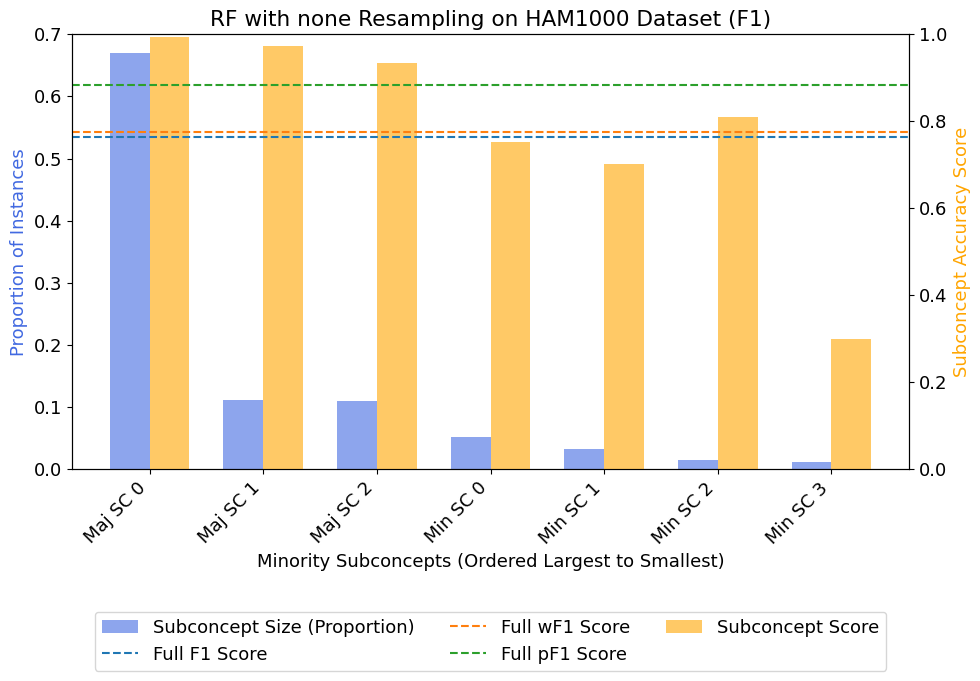}
    \end{minipage}
    \begin{minipage}{0.49\linewidth}
        \centering
        \appendixmedicalplot{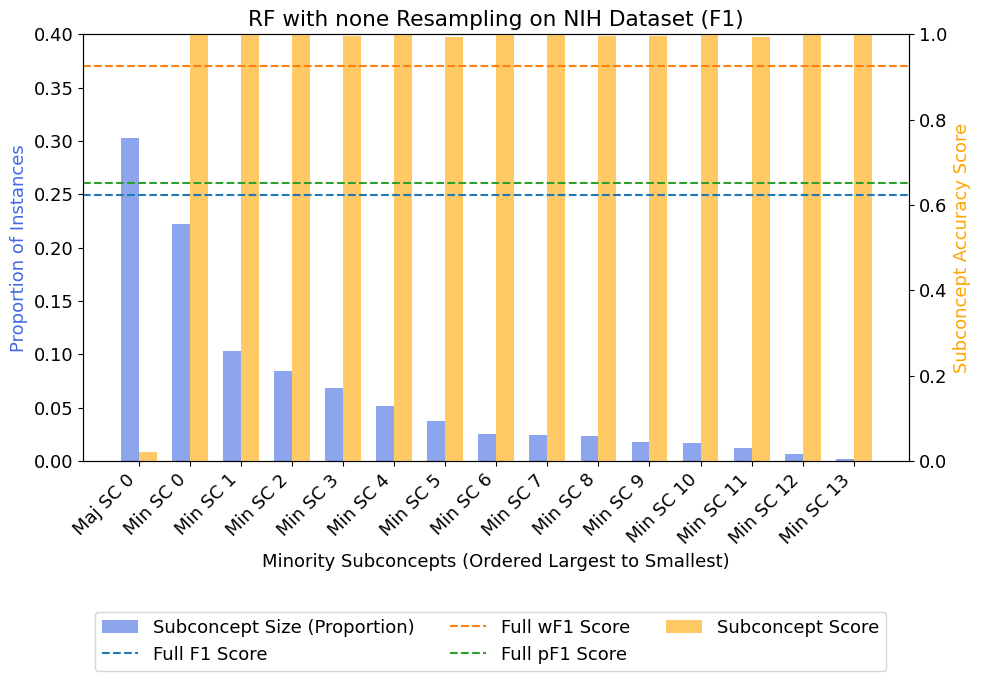}
    \end{minipage}

    \vspace{0.4em}

    \begin{minipage}{0.49\linewidth}
        \centering
        \appendixmedicalplot{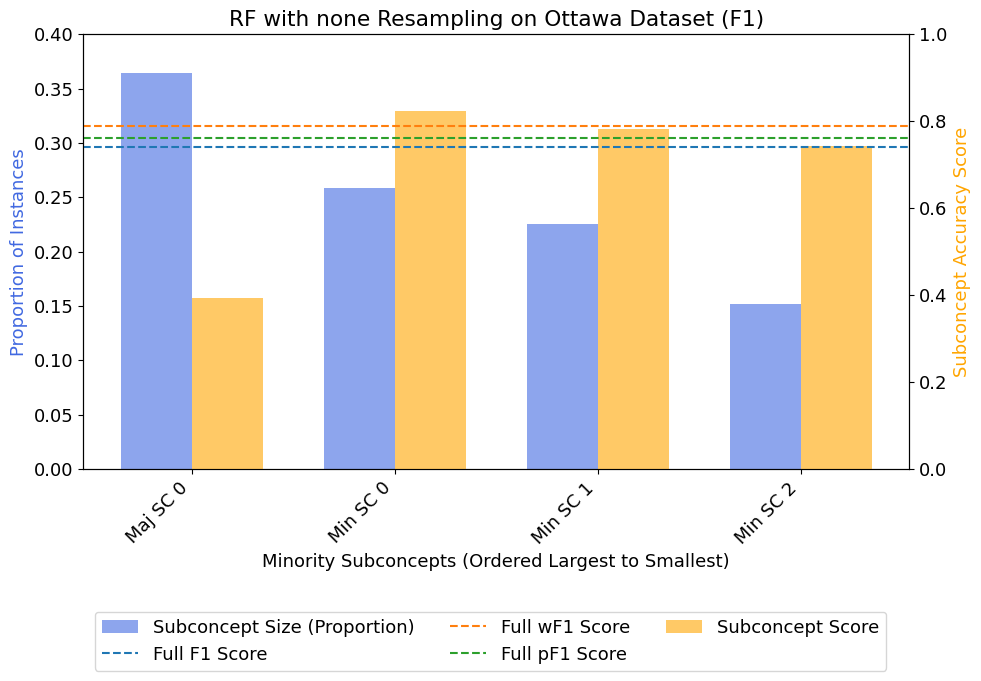}
    \end{minipage}
    \begin{minipage}{0.49\linewidth}
        \centering
        \appendixmedicalplot{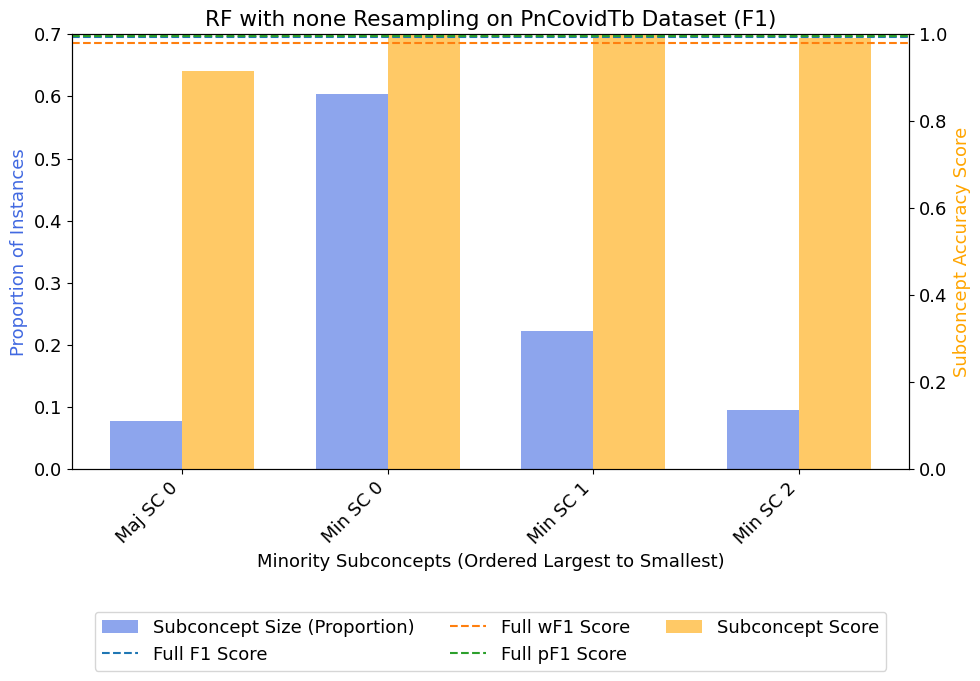}
    \end{minipage}
    \caption{F1 comparison plots on four medical datasets using embeddings followed by random forest classification. The interpretation of colors and lines is the same as in Figure \ref{fig:medical_emb_rf_ba}.}
    \label{fig:medical_emb_rf_f1}
\end{figure*}

\begin{figure*}[t]
    \centering
    \begin{minipage}{0.49\linewidth}
        \centering
        \appendixmedicalplot{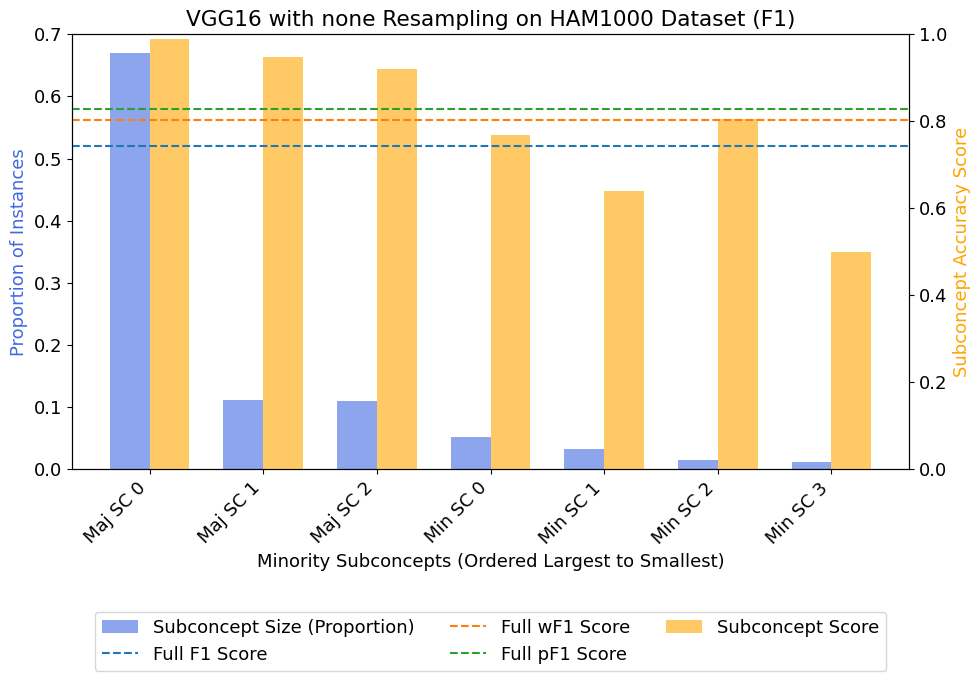}
    \end{minipage}
    \begin{minipage}{0.49\linewidth}
        \centering
        \appendixmedicalplot{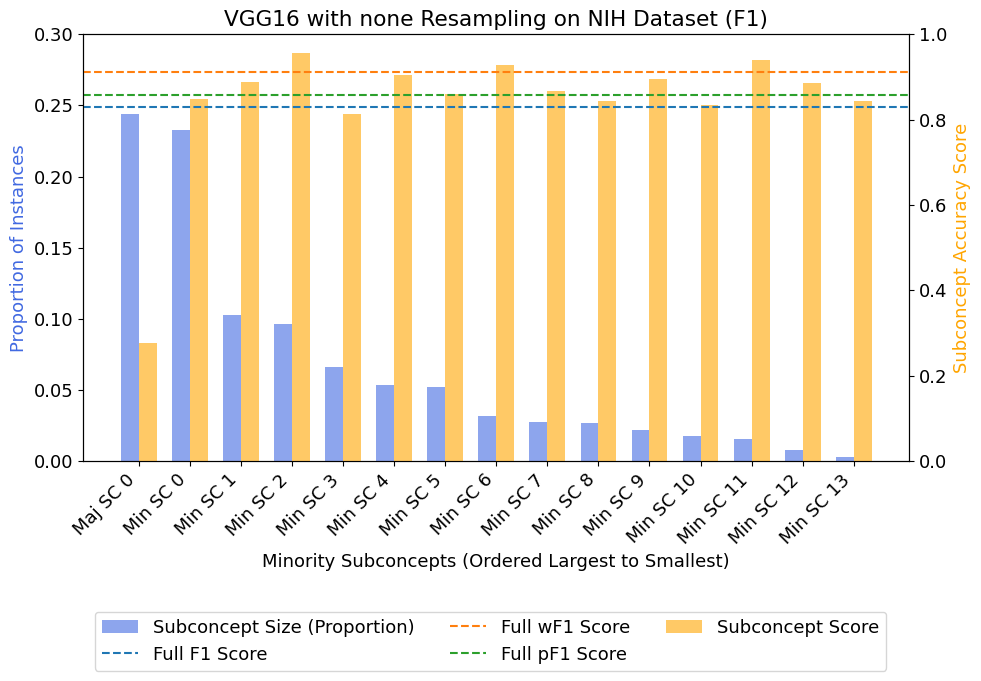}
    \end{minipage}

    \vspace{0.4em}

    \begin{minipage}{0.49\linewidth}
        \centering
        \appendixmedicalplot{figures/yaning/Ottawa_VGG16/results_parPriorPred/figures/comparison_plots/VGG16_none_f1_Ottawa.png}
    \end{minipage}
    \begin{minipage}{0.49\linewidth}
        \centering
        \appendixmedicalplot{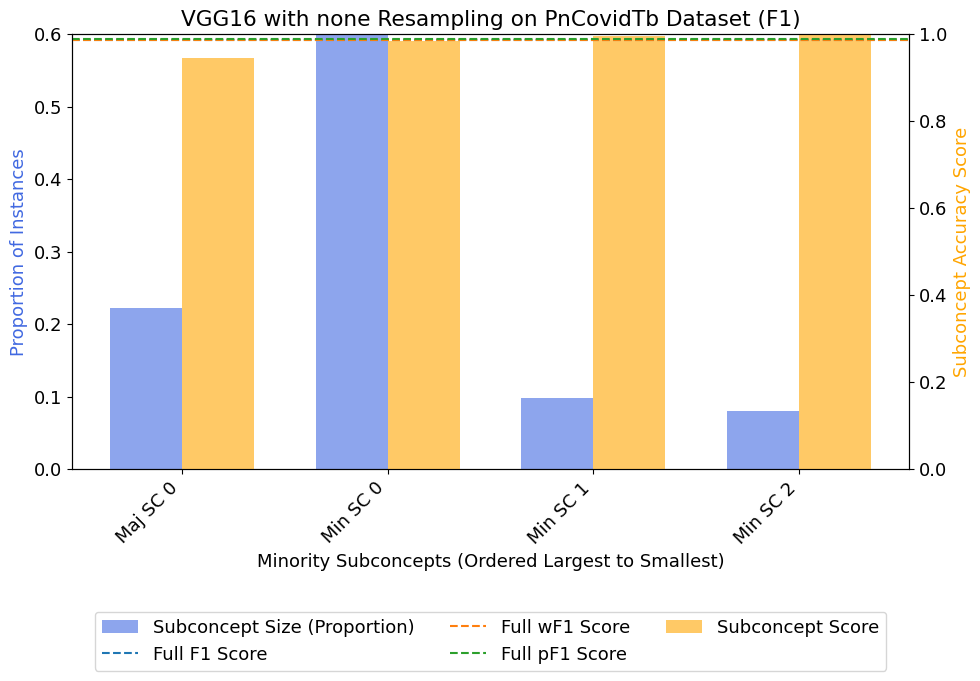}
    \end{minipage}
    \caption{F1 comparison plots on the same four medical datasets using direct VGG16 classification. The interpretation of colors and lines is the same as in Figure \ref{fig:medical_emb_rf_ba}.}
    \label{fig:medical_vgg_f1}
\end{figure*}

\end{document}